\documentclass{article}

\usepackage{PRIMEarxiv}

\usepackage[utf8]{inputenc}
\usepackage[T1]{fontenc}

\usepackage{hyperref}
\usepackage{url}

\usepackage{amssymb}
\usepackage{amsmath}
\usepackage{amsfonts}
\usepackage{bbm}

\usepackage{booktabs,tabularx,array,siunitx,ragged2e,makecell,multirow,enumitem}
\usepackage{longtable}

\usepackage{graphicx}
\usepackage{subcaption}

\usepackage{xcolor}
\usepackage{tcolorbox}

\usepackage{nicefrac}
\usepackage{microtype}
\usepackage{natbib}

\newcolumntype{d}{S[table-format=1.4]}
\newcolumntype{L}{>{\RaggedRight\arraybackslash}X}
\newcolumntype{i}{S[table-format=3.0]}
\newcolumntype{Y}{>{\raggedright\arraybackslash}p{2.3cm}}
\newcolumntype{N}{>{\centering\arraybackslash}p{0.7cm}}
\newcolumntype{Z}{>{\raggedright\arraybackslash}p{2cm}}
\newcolumntype{W}{>{\raggedright\arraybackslash}p{5cm}}
\newcolumntype{T}{>{\centering\arraybackslash}p{1.2cm}}
\newcolumntype{E}{>{\RaggedRight\arraybackslash}p{1.6cm}}

\usepackage{arydshln}

\title{ChatPlanner: A Large Language Model Framework for Personalized Public Transit Routing
}

\author{
  Tingting Yang\thanks{These authors contributed equally to this work.} \\
  \textsuperscript{1}School of Engineering and Materials Science, Queen Mary University of London \\
  Mile End Road, London, E1 4NS, United Kingdom \\
  \texttt{t.yang@qmul.ac.uk} \\
  \AND
  Chenhao Xue\footnotemark[1] \\
  \textsuperscript{2}Department of Engineering Science, University of Oxford \\
  Parks Road, Oxford, OX1 3PJ, United Kingdom \\
  \texttt{chenhao.xue@eng.ox.ac.uk} \\
  \And
  Jun Chen\thanks{Corresponding author} \\
  \textsuperscript{1}School of Engineering and Materials Science, Queen Mary University of London \\
  Mile End Road, London, E1 4NS, United Kingdom \\
  \texttt{jun.chen@qmul.ac.uk} \\
}

\begin{document}
\maketitle

\begin{abstract}
Personalized public transit routing in public transit systems remains challenging due to the difficulty of capturing and integrating diverse user preferences into routing algorithms. This paper presents ChatPlanner, a novel framework that leverages Large Language Models (LLMs) to enable preference-aware public transit routing. Our approach employs fine-tuned LLMs with Retrieval-Augmented Generation (RAG) to extract routing parameters and interpret conversationally expressed preferences from natural language queries as preference scores, subsequently integrating these preferences into the objective function of a public transit routing algorithm. This study designs preference-aware datasets incorporating eight personas and five contexts to establish scoring standards for both fine-tuning and RAG. This work conducted four experiments to validate the solutions' feasibility, extraction of routing information and preferences, solution set quality and completeness, and latency and computational tractability. Results demonstrate that ChatPlanner generates feasible solutions reliably. Fine-tuning enforces the required output structure and learns general preference patterns, while RAG provides query-specific context to resolve imprecise or conversational expressions and calibrate continuous scores. The combination of both achieves the highest accuracy in routing information extraction and rubric-consistent user preference interpretation. Results based on selected case studies show that by capturing user conversationally expressed preferences, ChatPlanner identifies preference-relevant solutions across different dimensions that existing route planners overlook, generating more route alternatives. The latency evaluation confirms that the framework is computationally tractable. This research establishes a new paradigm for integrating natural language understanding into transportation optimization.

\end{abstract}



\section{Introduction}
\label{sec:introduction2}

Public transit, as defined by the Federal Transit Administration, is transportation by bus, rail, or other conveyance, either publicly or privately owned, that provides general or special service to the public on a regular and continuing basis \citep{Daganzo2019}. These systems are characterized by large-capacity vehicles operating on predetermined routes and timetables. They serve as the foundation of urban mobility in major cities worldwide, particularly in areas with high population density \citep{horcher2021review}. Public transit routing identifies feasible passenger paths across existing transit networks. Public transit routing algorithms provide decision support for route planner apps, for example, Google Maps, Apple Maps, and Citymapper. However, most commercial journey planners still primarily optimize for travel time and the number of transfers as fixed objectives. These planners do not interpret users' preferences directly and translate those into extra objectives dynamically \citep{applemaps2025, googlemaps2025,citymapper2025}. 

Traveler preferences are heterogeneous across time sensitivity, trip purpose, accessibility, affordability, and perceived safety. Commuters typically prioritize time efficiency \citep{mackie2001value}, leisure travelers often trade time for itinerary quality \citep{zhou2024tourists}, and riders with limited mobility require step-free access \citep{ceccato2020measure}. Low-income travelers, non-drivers, children, and elderly adults frequently rely on public transit for affordability, safety, and convenience \citep{fan2004optimal}, while safety concerns, particularly among women, children, and students, also shape route choice \citep{mayorlondon2023violent}. Because public transit is essential to daily life for many groups, system performance has tangible consequences for well-being \citep{heaps2021public,molner2023policy}. Incorporating user preferences into routing algorithms is therefore critical to serving diverse needs and improving the overall transit experience.

When specific preferences are captured through the user interfaces of those commercial journey planners, they are often used as post-hoc filters. That is, the solution set and the process of finding solutions remain unchanged. The routes are simply re-ranked by preferences. Public transit routing problems with multiple conflicting objectives seek to find Pareto optimal journeys for travelers \citep{delling2015round, fan2010metaheuristic}. However, considering many conflicting preferences as various objective functions in the public transit routing algorithm may cause high computational cost, and lead to information clutter as more than necessary solutions are presented to the end users.

Apart from the limitations of fixed objectives, a more fundamental issue lies in how preferences are captured in the first place. A structured preference interface with sliders or ranking widgets could let travelers rate criteria such as accessibility or safety directly. However, research in public transit routing shows that it is difficult for users to accurately express their preferences for each criterion through seemingly more guaranteed inputs \citep{wang2021personalized, huang2024personalized}. Such an interface faces an inherent trade-off: simplified but more guaranteed methods require less cognitive effort but are less expressive, while more expressive methods demand greater cognitive effort and reduce usability \citep{pommeranz2012designing}. However, even when the preference dimensions are low, asking users to assign numerical importance scores creates a cognitive mismatch. Travelers naturally think in terms of situational context (e.g., ``I have a stroller and it is rush hour'') rather than abstract ratings of preferences on a scale \citep{kostric2024generating}.

Given the above circumstance, this study contemplates three research questions: 
(i) What constitutes an effective, systematic methodology for formulating a useful framework to identify travelers’ preferences?
(ii) How does the proposed framework better understand traveler requests across different preferences?
(iii) How does the proposed work perform in both theoretical and practical perspectives?

To address the above challenges, ChatPlanner, a Large Language Model (LLM) Framework for Personalized Public Transit Routing, is proposed. Within this context, \textit{Chat} corresponds to the conversational interaction and natural language handling functionalities provided by LLM. \textit{Planner} applies a multi-criterion round-based public transit routing (MC-RAPTOR) algorithm \citep{delling2015round}. Practically, ChatPlanner provides a flexible conversational interface that interprets user's conversationally expressed preferences and returns diverse, feasible route sets. Methodologically, a fine-tuned transit-domain LLM translates natural-language queries into journey requests and user preferences based on four extra criteria in addition to travel time and number of transfers: (i) accessibility level, (ii) crowding level, (iii) safety level, and (iv) sightseeing level. The LLM estimates the importance of these criteria, allowing the framework to select the most relevant ones for the traveler. The route planner incorporates these preferences from the traveler into the multi-objective search and selects potentially different objective functions during routing. These preference-aware objectives thereby yield distinct Pareto fronts over candidate routes, combining personalized user preferences with the transit routing algorithm.

The contribution of this framework is that, first, the state-of-the-art LLM models' performances in generating public transit routing solutions are investigated, providing insights into the feasibility, completeness, and correctness. Second, this paper introduces ChatPlanner, a framework that integrates the LLM with a public transit routing algorithm with a preference-aware dataset. This work designs the datasets for both fine-tuning and RAG with an input-output structure and incorporates preference information through eight distinct personas and five contexts to establish preference scoring standards. Third, this study evaluates the proposed RAG and fine-tuned LLM under the proposed framework for extracting routing information and interpreting user conversationally expressed preferences, compared to general purpose base LLM models. Evaluation demonstrates that fine-tuning is essential for accurate parsing of origin, destination, and requested arrival time, while the combination of fine-tuning and RAG best enhances rubric-consistent preference interpretation accuracy. Especially, the RAG's contextual examples ensure consistent and accurate preference scoring, aligning with the designed scoring standards. In addition, the framework's end-to-end latency and computational tractability are evaluated for real-world deployment. Finally, this study conducted case studies to confirm that by systematically capturing users' preference expressions as preference scores, the framework generates more route alternatives overlooked by existing route planners.

The rest of this article is organized as follows. Section 2 describes the related work on classic public transit routing algorithms, multi-objective routing algorithms, personalized route recommendation systems, LLM in transportation, and LLM in public transit routing. Section 3 provides a detailed presentation of the proposed ChatPlanner framework. Section 4 presents the experiments conducted to test solution feasibility, parsing performance, solution diversity, and the system's latency and computational tractability, and discusses the major findings and their academic and industrial implications. Section 5 draws the conclusion, and Section 6 outlines future research directions.

\section{Literature Review}
\label{sec:literature2}

A public transit system is generally referred to as public transport and mass transit. When it comes to public transit, bus, railway, and airline transit network design play a key role in optimization \citep{kepaptsoglou2009transit,cipriani2012transit, liang2025novel}. \citet{kepaptsoglou2009transit} systematically reviews research on the Transit Route Network Design Problem, addressing the challenges of efficiently designing public transportation networks that enhance mobility while reducing congestion and pollution. \citet{liang2025novel} propose a novel multi-objective evolutionary algorithm based on objective space decomposition to solve the transit network design and frequency-setting problem. The network design problem determines the stations that the bus, subway, or railway needs to go through at a specific frequency. By contrast, the public transit routing aims to select the stations that travelers need to go through from the origin to the destination, given the scheduled timetable. 

Early approaches to public transit routing adapt classical shortest path algorithms, particularly Dijkstra's algorithm, by modeling timetables as either time-expanded or time-dependent graphs \citep{pyrga2008efficient,disser2008multi}.
\citet{delling2015round} introduce Round-based Public Transit Optimized Router (RAPTOR), which is achieved by examining routes that have been marked and incrementally exploring connected paths in a round-by-round manner, identifying the earliest trip on that line that can be boarded. Unlike previous Dijkstra-based graph approaches, RAPTOR operates in rounds, processing each route at most once per round, making it simpler and faster. MC-RAPTOR is a multiple-criteria Round-based Public Transit Optimized Router, a multiple-criteria extension of RAPTOR for computing Pareto-optimal journeys in public transit networks that minimizes arrival time, number of transfers, and any extra criteria in a round-based manner. In each iteration, it determines the earliest arrival times for journeys with $n-1$ transfers, where $n$ corresponds to the current iteration. The multiple objective values are saved in different labels. Not all public transit routing problems are NP-Hard, but they become NP-Hard when multiple objectives need to be considered.

Regarding the traveler's preference, \citet{jakob2014personalized} presents an advanced journey planner that combines all types of urban mobility services while personalizing recommendations based on each user's past travel choices and preferences.
\citet{ceder2019personalized} considers multiple traveler preferences, including time, cost, and convenience, and performs multiple searches across different criteria to generate alternative routes, with each route optimized for different attribute combinations to provide personalized options. \citet{zografos2010identifying} identify traveler preferences by conducting surveys that capture travelers' needs, which are then analyzed to determine the requirements for personalized journey planning services.
\citet{ludwig2009recommendation} propose ROSE, a mobile application that combines event recommendation with public transportation navigation, using the A* algorithm to incorporate multi-criteria user preferences for optimal public transit routing. \citet{7580779} develop a personalized route recommendation system that identifies commuter preferences through a participatory sensing Android application, which collects convenience feedback from users during their public transit journeys, and stores this feedback in a knowledge base to provide personalized route recommendations.

In recent years, a new paradigm has been proposed to create greener, safer, and more inclusive future transportation \citep{10154168,shamsuddoha2025review}. This new paradigm is reflected as transportation 5.0, which employs emerging technologies such as Internet of Things, artificial intelligence (AI), and data analytics \citep{10154168,mobipar_mobility_5_0_2024_online}. An LLM is an AI model that learns language patterns by training on large datasets. These models are built to understand and generate human language. Although language models have been developed for decades, transfer learning is what makes foundation models possible, and scale is what makes them powerful in recent years \citep{weizenbaum1966eliza,bommasani2021opportunities}. The introduction of the self-attention mechanism enabled the transformer architecture to capture long-range dependencies more effectively \citep{vaswani2017attention}. LLMs possess remarkable capabilities, producing text that can match human writing in both quality and natural flow. These powerful AI systems have rapidly become integrated into people's everyday lives \citep{rapp2025people}.

The emergence of LLMs in transportation can be seen in recent years. \citet{cheng2025llm} propose LLM-TFP, which integrates LLMs with spatio-temporal features through a tokenizer that captures timestep, time-of-day, and spatial embeddings for urban traffic flow prediction. \citet{rong2024edge} propose an architecture that integrates LLMs with spatio-temporal features for large-scale traffic flow prediction in intelligent autonomous transport systems. \citet{wang2025agentic} introduces LLMTraveler, an LLM-empowered agent with a memory system that learns from past experiences to simulate human route choice behavior in transportation networks. \citet{zhao2025safetraffic} integrates multi-modal crash data, including driver, vehicle, and environment information, into a textualized dataset of prompts to fine-tune an LLM for crash prediction, and provides event-level and conditional risk analysis for identifying critical features and guiding targeted interventions.

For the LLM in public transit routing problem, \citet{jonnala2024usinglargelanguagemodels} use LLMs to handle questions about the General Transit Feed Specification (GTFS) data, which is a data format that contains a series of .txt files for public transit. Questions such as ``What is the meaning of the accessibility file?'' target the understanding of the GTFS data format. Furthermore, there is a gap between analyzing the GTFS data and providing journey solutions. \citet{wang2024leveraging} design an LLM-based trip advisor application to provide trip suggestions. The key inputs are the Origin-Destination pair (OD), travel time, and user preference. This study uses LLMs to generate paths directly based on information queried from their database, without using any public transit routing algorithms. \citet{fang2024travellmplannewpublic} apply LLMs in public transit routing for network disruption. The LLM works as both planner and summarizer. The LLM planner analyzes the subway map for connections between stops and routes through stops. It can identify nearby stations and possible routes as alternatives during disruption. The LLM summarizer organizes the content about possible routes and stops. However, these suggestions do not include timetable information for suggested routes and stops. \citet{fang2024travellmplannewpublic} show that it is difficult to conduct the experiments since there is no benchmark model for LLM in public transit routing. Also, the challenges of LLM in public transit routing lie in its lack of transportation-specific domain knowledge, public transit data, and the accuracy of its answers.

Despite recent advances in applying LLMs to transportation problems, significant gaps remain in personalized public transit routing. Existing personalized routing approaches rely heavily on historical data collection through surveys, past travel records, or participatory sensing applications to learn user preferences. These methods assume static preference patterns and require substantial user data before providing personalized recommendations. They fail to address the reality that travelers' requests are dynamic and change daily based on varying contexts, needs, and circumstances. Recent LLM-based transit applications have limitations. Some focus only on interpreting GTFS data structures without providing actual journey solutions. Others employ routes recorded in the past directly without using a routing algorithm to find the most suitable routes according to the current dynamic real-world environment. Most lack the ability to dynamically interpret and adapt to user preferences expressed in natural language during the planning process.

Therefore, this work addresses the gap by proposing an LLM-based framework that enables intelligent, personalized preferences by integrating LLM with MC-RAPTOR to find solutions based on real timetable data. The proposed framework supports conversational preference refinement throughout the journey planning process and allows public transit routing to be personalized, intelligent, and adaptive to the dynamic nature of daily travel requests. Within the large language models for transportation taxonomy~\citep{nie2025exploring}, ChatPlanner operates simultaneously as an information processor, interpreting natural language preference expressions, and a decision facilitator, integrating LLM reasoning with MC-RAPTOR to produce timetable-grounded journey plans.

\section{Methodology}
\label{sec:methodology}


This section explains the ChatPlanner framework and its core components. Section~\ref{sec:chatplanner_framework} presents the overall framework architecture, which focuses on the information transmission across each node. Furthermore, two core components are explained in detail in Sections~\ref{sec:parse_llm_rag} and \ref{sec:mc-raptor_methodology}, including the fine-tuned LLM parser with RAG for extracting routing information and predicting preference scores from traveler requests in Node 1, and the process of integrating the multi-criteria public transit algorithm in Node 4.

\subsection{ChatPlanner Agent Framework}\label{sec:chatplanner_framework}

ChatPlanner is structured as a directed graph of seven nodes, implemented via the LangGraph software library \citep{langgraph2024}.

\begin{figure}[t]
\centering
\includegraphics[width=1\textwidth]{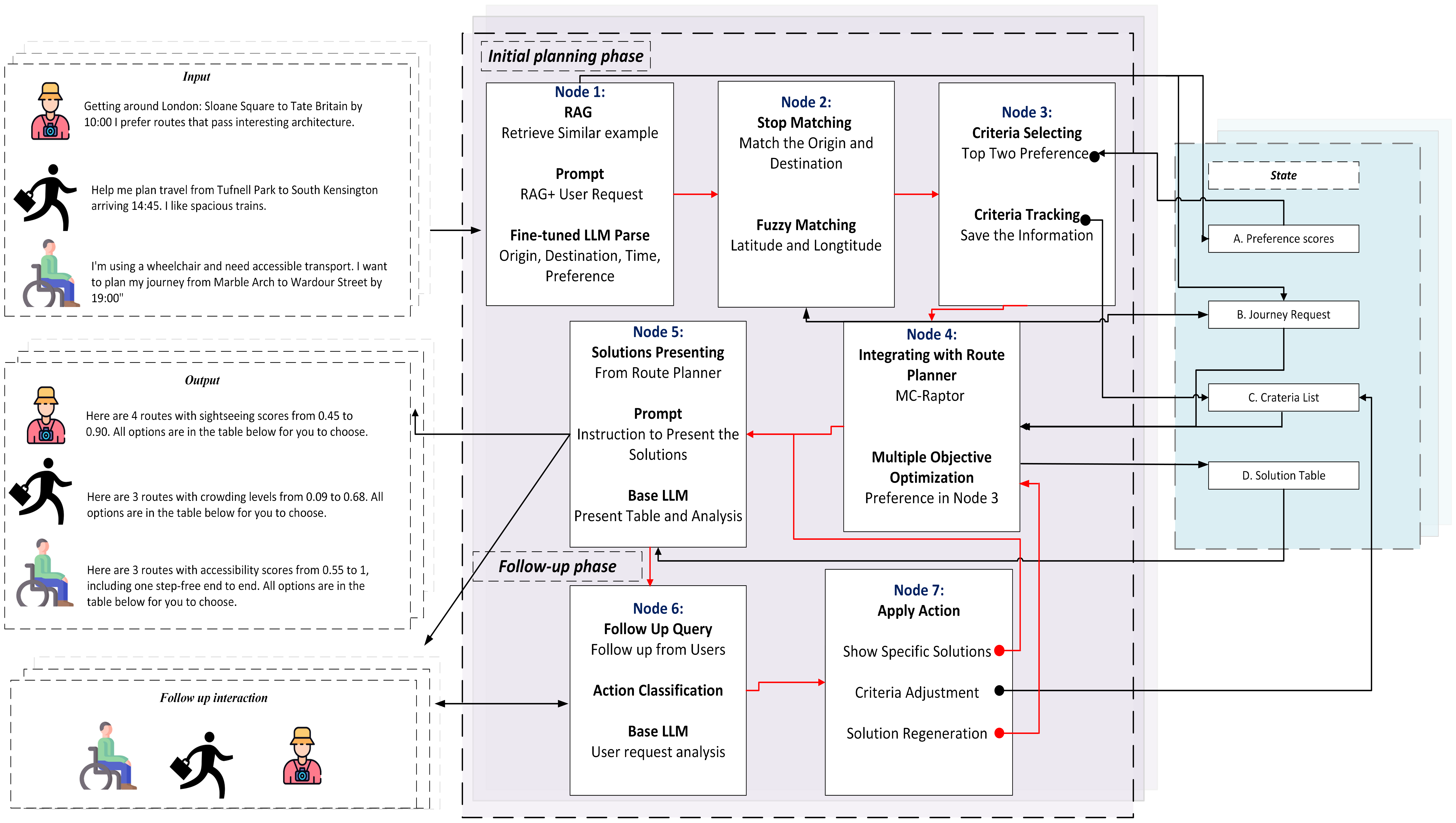}
\caption{Overall framework of the LangGraph for ChatPlanner}\label{fig:agent_architecture}
\end{figure}

The LangGraph architecture comprises three core elements: nodes, edges, and a shared data structure State, as illustrated in Figure \ref{fig:agent_architecture}. State stores structured information throughout execution, including Preference Scores, Journey Request, Criteria List, and Solution Table. Nodes represent discrete actions. Each node operates on State through a read-modify-write cycle: it reads the structured information, performs its designated action, and updates State with results. Edges define conditional transitions between nodes, determining control flows. 



\subsubsection{Initial Planning Phase}\label{sec:initial_phase}

\textit{Node 1} contains three actions: RAG, Prompt, and Fine-tuned LLM parse. It processes the user's free-text input and records the requested origin, destination, and arrival time. The input may also include the user's preferences and intentions. First, the semantic similarity is computed between the user's input query and the query instances stored in the RAG dataset. Second, the five most similar RAG instances are retrieved and injected into the Prompt together with the user's input query. The Prompt instructs the fine-tuned LLM to parse the correct request of the user's information, including the origin, destination, requested arrival time, and preferences. The parsed information from Node 1 updates the Preference Scores and Journey Request in the State.


\textit{Node 2.} The origin and destination from the Journey Request, parsed by the fine-tuned LLM, may not match the standard format required by the Route Planner in Node 4. The parsed origin and destination are mapped to standard names that appear in the available timetables for the Route Planner to carry out the search. The mapping is supported by both the string matching check and a further geographic matching check using latitude and longitude to avoid misinterpretation. The geographic matching check using the geopy package \citep{geopy} improves the mapping accuracy, even when the user uses conventions or abbreviations for stations. The accurate origin and destination are written to the State.

\textit{Node 3.} The Preference Scores from State contain four criteria: accessibility, crowding, safety, and sightseeing. Each criterion is parsed with a score, from 0 to 1, in Node 1 by the fine-tuned LLM. These preference scores require further processing in Node 3 for active preference selection. To accurately identify actively expressed user preferences and maintain computational tractability during route search, Node 3 implements a two-step preference selection process. First, criteria with scores below $T=0.6$ are filtered out. This threshold $T$ was determined empirically through preliminary experiments. By evaluating 30 semantically neutral journey requests (i.e., requests containing only origin, destination, and time information), it is observed that the average predicted scores naturally hover around $0.5$ rather than absolute zero (detailed results of this empirical test are provided in \ref{apx:threshold_justification}). Therefore, $0.6$ is established as the minimum activation threshold. Second, from the retained set, a maximum of $K=2$ criteria with the highest scores are selected as active user preferences for the initial planning phase. A sensitivity analysis on the value of $T$ and $K$ is conducted and presented in Appendix B. Because the MC-RAPTOR algorithm inherently optimizes travel time and the number of transfers, adding $K=2$ user preferences creates a four-objective Pareto optimization problem. To guarantee that moderately important preferences will not be discarded, and to ensure this selection policy does not systematically suppress solutions or cause permanent information loss, ChatPlanner leverages its conversational architecture. The parameter $K$ can be adjusted, and any specific preferences filtered out in Node 3 can be reselected during the follow-up conversation phase (Node 6) if the user is unsatisfied with the initial solutions. The selection is saved to the Criteria List in State. Actions in Nodes 1, 2, and 3 work together to translate the user's expressed preferences into active routing criteria.

\textit{Node 4} reads the Journey Request and the Criteria List from State, including the canonical origin and destination, the requested arrival time, and the active user preferences as inputs. Then these inputs are passed to MC-RAPTOR, with the active user preferences determining the objectives that drive search. The solutions provided by MC-RAPTOR are saved in the Solution Table in State.
More information about MC-RAPTOR is explained in Section~\ref{sec:mc-raptor_methodology}.



\textit{Node 5} is designed to present the solutions, which simplifies the Solution Table essential information. The original table contains case metadata, algorithm parameters, intermediate metrics, and final criteria values. The simplified table retains only the solution ID and six solution criteria: travel time, number of transfers, accessibility, crowding, safety, and sightseeing. This simplification reduces users' effort in comprehending solutions. A base LLM then presents this simplified table together with a brief summary. The summary does not rank the solutions or single out any solution; the complete solution set is always presented, and the final route choice is left to the user.


\subsubsection{Follow-up Phase}

Nodes 6 and 7 work together to handle the follow-up phase. Node 6 interprets the follow-up request from users, leveraging a base LLM for conversational interaction. The main responsibility of Node 6 is to decide which actions to take in Node 7. The actions include: (i) Show Specific Solutions, (ii) Criteria Adjustment, and (iii) Solution Regeneration. Show Specific Solutions is executed when the user requests to provide more details for the existing solutions from the initial planning phase. Criteria Adjustment happens when Node 6 detects that the user is particularly emphasizing one of the four criteria. It triggers the solutions re-ranking process based on the emphasized criterion. Solution Regeneration initiates re-planning based on the follow-up request by calling the route planner in Node 4. When the users provide an extra criterion and are not satisfied with the current solutions, the regeneration starts.


\subsection{Fine-tuning LLM Parse and RAG}\label{sec:parse_llm_rag}

In the LangGraph architecture, Nodes 1, 5, and 6 all invoke an LLM, but each node assigns the LLM a distinct role. Nodes 5 and 6 rely on the base LLM to present solutions and to classify user intended follow-up actions. These do not require fine-tuning. By contrast, Node 1 employs a fine-tuned LLM for parsing the origin, destination, and requested arrival time from the user’s query and to interpret the user’s stated travel preferences.

\subsubsection{Fine-tuned Data Preparation}
Two established datasets serve as the foundation for fine-tuning: the Multi-Domain Wizard-of-Oz (MultiWOZ) and Schema-Guided Dialogue (SGD) datasets Buses domain. MultiWOZ was chosen because it contains over 10,000 human-to-human dialogues, including many in train and taxi transportation contexts \citep{budzianowski2018multiwoz}. SGD was selected because it contains 3135 annotated dialogues for bus services \citep{rastogi2020towards}. In MultiWOZ, stop names are real UK city names and railway stations, while in SGD, they are primarily US city names. This geographic diversity in the training data provides varied location naming patterns.

To prepare the fine-tuning dataset, the MultiWOZ dataset is filtered to retain only transportation-related dialogues. Because the focus is on urban public transit in this work rather than intercity trains or taxis, this study replaces train and taxi terms with urban transit terms in both datasets. For example, “find me a train” becomes “find me a route,” and “book a taxi” becomes “plan a route,” to ensure the terminology is consistent with urban transit. Since MC-RAPTOR searches timetables backward from a requested arrival time, this study uses an ``arrive by'' convention rather than “depart at” throughout the dataset. While the current system does not implement departure time logic, the same data processing approach could be applied by converting ``arrive by'' to ``depart at'' convention and integrating forward search functionality in MC-RAPTOR.

\paragraph{Preference Data Augmentation}

\begin{table}[htbp]
\footnotesize
\centering
\caption{Persona Types, Preference Weights, and Examples. \textit{Note: Personas are not included in the LLM's inputs and predicted outputs during both training and generation.}}
\label{tab:personas}
\renewcommand{\arraystretch}{1.15}
\begin{tabularx}{\textwidth}{Y N N N N X}
\toprule
{\footnotesize\textbf{Persona}} &
\multicolumn{1}{c}{\makecell{\footnotesize\textbf{Access-}\\\textbf{ibility}}} &
\multicolumn{1}{c}{\makecell{\footnotesize\textbf{Crowd-}\\\textbf{edness}}} &
\multicolumn{1}{c}{\footnotesize\textbf{Safety}} &
\multicolumn{1}{c}{\makecell{\footnotesize\textbf{Sight-}\\\textbf{seeing}}} &
{\footnotesize\textbf{Example}} \\
\midrule
Sightseeing Tourist   & 0.4  & 0.3  & 0.7  & 0.9  & I'd love to see landmarks on the way \\
Mobility Impaired     & 0.95 & 0.8  & 0.8  & 0.2  & I don't want stairs or gaps \\
Safety Conscious       & 0.4  & 0.6  & 0.95 & 0.2  & I want to feel secure in my journey \\
Crowd Averse          & 0.4  & 0.9  & 0.6  & 0.3  & I prefer quieter, less busy routes \\
Daily Commuter        & 0.6  & 0.7  & 0.7  & 0.2  & I go to the office \\
Family with Children  & 0.8  & 0.8  & 0.9  & 0.6  & I'm traveling with children \\
Senior Citizen        & 0.9  & 0.85 & 0.9  & 0.5  & I am elderly; I want less crowded routes \\
Flexible Travel       & 0.5  & 0.3  & 0.6  & 0.7  & I'm flexible with conditions \\
\bottomrule
\end{tabularx}
\end{table}

\begin{table}[htbp]
\footnotesize
\centering
\caption{Context Types, Preference Weights, and Examples. \textit{Note: Contexts are not included in the LLM's inputs and predicted outputs during both training and generation.}}
\label{tab:context-adjust}
\renewcommand{\arraystretch}{1.15}
\begin{tabularx}{\textwidth}{Z N N N N X}
\toprule
{\footnotesize\textbf{Context}} &
\multicolumn{1}{c}{\makecell{\footnotesize\textbf{Access-}\\\textbf{ibility}}} &
\multicolumn{1}{c}{\makecell{\footnotesize\textbf{Crowd-}\\\textbf{edness}}} &
\multicolumn{1}{c}{\footnotesize\textbf{Safety}} &
\multicolumn{1}{c}{\makecell{\footnotesize\textbf{Sight-}\\\textbf{seeing}}} &
{\footnotesize\textbf{Example}} \\
\midrule
Rush Hour            & 0.6 & 0.8 & 0.5 & 0.5 & during rush hour; in the morning rush; when it's busy \\
Late Night   & 0.5 & 0.3 & 0.9 & 0.2 & late at night; after dark; in the evening \\
Weekend      & 0.5 & 0.6 & 0.7 & 0.8 & on the weekend; for a weekend trip; on Saturday/Sunday \\
Tourist Area & 0.5 & 0.6 & 0.7 & 0.9 & in the tourist district; near the main attractions \\
Bad Weather  & 0.8 & 0.7 & 0.5 & 0.3 & when it's raining; during bad weather \\
\bottomrule
\end{tabularx}
\end{table}


Our fine-tuning approach follows a supervised learning setting, where the model learns to map traveler queries in the form of natural language to structured JSON outputs containing origin, destination, requested arrival time, and preference scores. In supervised fine-tuning, the model requires labeled training examples with input-output pairs that demonstrate the desired behavior. Since existing traveler query dialogue datasets lack preference annotations, this study conducts systematic preference data augmentation guided by established travel behavior research to inject preference information into journey requests and create the necessary ground truth labels for our training dataset.

An augmentation system is designed based on eight user personas and five context types (Tables~\ref{tab:personas}, \ref{tab:context-adjust}). The eight personas defined include mobility impaired users \citep{ceccato2020measure, bezyak2017public}, safety-conscious travelers \citep{loukaitou2014fear, delbosc2012modelling}, crowd-averse passengers \citep{roncoli2023estimating, tirachini2013crowding}, sightseeing tourists \citep{zhou2024tourists, de2016travel}, daily commuters \citep{cascetta2014hedonic}, families with children \citep{delbosc2012modelling, ceccato2020measure}, senior citizens \citep{ncat2024barriers}, and flexible travelers \citep{cascetta2014hedonic}.

Preference weights are assigned to each persona across four dimensions: accessibility, crowdedness, safety, and sightseeing, with the score of preferences referencing the above literature. For example, high accessibility preferences are assigned to mobility-impaired 
users with accessibility scores set to 0.95 in the dataset, consistent with 
findings on accessibility measurement \citep{ceccato2020measure}. Similarly, 
high sightseeing scores (0.9) are assigned to tourist personas, following 
\citet{zhou2024tourists}. Contextual modifiers are designed to assign preference scores for situational factors such as rush hour, late night, weekend, tourist area, and bad weather conditions, where score magnitudes are guided by \citep{loukaitou2014fear, cascetta2014hedonic}. These assignments serve as the ground truth labels for the training dataset, though alternative scoring schemes could be developed for different preference frameworks.

The augmentation process randomly selects personas or contexts for each training example, injects appropriate preference language into the instruction, and generates the corresponding preference scores in structured JSON output, creating aligned input-output pairs for supervised learning. The JSON structure ensures the fine-tuned model produces consistently parseable output.

\subsubsection{Fine-tune}
Fine-tuning adapts pre-trained language models to domain-specific tasks by optimizing model parameters $\theta$ to minimize the loss function.

\begin{equation}
\mathcal{L}(\theta) = \frac{1}{N} \sum_{i=1}^{N} \ell(f_\theta(x_i), y_i)
\end{equation}
where $f_\theta$ represents the language model, $(x_i, y_i)$ are input-output pairs from the fine-tuned dataset, and $\ell$ is the loss function. 

Parameter Efficient Fine-Tuning (PEFT) \citep{han2024parameter} is employed to reduce computational costs. Specifically, Low-Rank Adaptation (LoRA) \citep{hu2022lora} is used, which avoids updating all model parameters by learning low-rank small adjustment matrices. For each linear layer of weight matrix $W_0 \in \mathbb{R}^{d \times k}$ in the original model, LoRA introduces two smaller matrices $B \in \mathbb{R}^{d \times r}$ and $A \in \mathbb{R}^{r \times k}$, where $r$ is the rank and $r \ll \min(d,k)$. The adapted transformation becomes:

\begin{equation}
h = W_0 x + B A x
\end{equation}

The original weights $W_0$ remain frozen, and only $B$ and $A$ are trained. This reduces trainable parameters from $d \times k$ to $r \times (d + k)$, significantly lowering computational requirements.

LoRA is applied to the LLaMA-3-8B-Instruct and Qwen2.5-7B-Instruct models with rank $r=16$ and scaling factor $\alpha=32$. The method targets the attention and feed-forward layers of these models. Training uses 4-bit quantization for memory efficiency and runs for 2 epochs with learning rate $1.5 \times 10^{-4}$. The objective is to train the model to parse natural language transportation queries into structured JSON outputs containing origin, destination, arrival time, and preference scores. Please note that the persona and context are not used as labels to form the training dataset; the LLM is trained to perform direct regression from the raw natural language request to predict the four continuous preference scores.

\subsubsection{Retrieval-Augmented Generation (RAG)}

RAG \citep{lewis2020retrieval} enhances language model performance by incorporating relevant external knowledge during inference. While fine-tuning updates the model's global parameters to learn the required JSON output structure and general preference scoring patterns, RAG provides instance-specific context; by retrieving similar examples as local reference points to guide the fine-tuned model's output generation, it helps the model resolve users' imprecise and scenario-specific expressions and addresses scenario-specific preference interpretation.

The retrieval process identifies semantically similar examples using text embeddings. An embedding is a mathematical representation that converts text into a numerical vector, where texts with similar meanings produce vectors that are close together in vector space. The BAAI/bge-small-en-v1.5 embedding model \citep{xiao2024c} is used to compute 384-dimensional vectors for query instance $q_i$ in our RAG dataset, denoted as $\mathbf{e}_i$. Similarly, the user query $q$ is converted to a query embedding $\mathbf{e}_q$.

The semantic similarity between a user query and a query instance stored in the RAG dataset is measured using cosine similarity:

\begin{equation}
\text{sim}(q, q_i) = \frac{\mathbf{e}_q \cdot \mathbf{e}_i}{||\mathbf{e}_q||_2 \cdot ||\mathbf{e}_i||_2}
\end{equation}
where $\mathbf{e}_q \cdot \mathbf{e}_i$ is the dot product and $||\cdot||_2$ denotes the L2 norm. This metric ranges from -1 to 1, with higher values indicating greater semantic similarity.

For each user query, the top $k=5$ most similar queries are retrieved from the RAG dataset. These retrieved examples are then formatted as demonstration examples in the prompt together with the current user query. This approach, known as in-context learning, provides the fine-tuned model with concrete examples showing how similar queries were previously mapped to structured preference scores. The retrieved examples help the model maintain consistency with our designed preference score assignments based on personas and contexts.

The retrieval library is constructed using the same training dataset generated through the preference data augmentation process. Consequently, highly similar patterns exist between the training set and the RAG corpus. It is important to note that the retrieved examples contain only the raw natural language request and the corresponding origin, destination, arrival time, and preference scores; they do not explicitly contain the persona or context labels used during preference data augmentation. Furthermore, the retrieval library is strictly disjoint from the human-written evaluation test set used in Section 4.2, ensuring that the model is tested on its ability to generalize to out-of-distribution human language rather than simply memorizing training distributions.

\subsection{Multi-Criteria Route Planning Algorithm}\label{sec:mc-raptor_methodology}

The Multi-criteria Round-Based Public Transit Routing (MC-RAPTOR) algorithm is used as the default solver in Node 4. The original MC-RAPTOR algorithm uses timetable data to optimize travel time and the number of transfers. MC-RAPTOR runs backward in the timetable from the destination, marking routes and stops that can reach the destination. The search proceeds in rounds $N$; stops marked in the previous round seed the expansion in the next round. The number of transfers is equal to $N-1$. Per-stop labels store objective values, including travel time labels $L_1$, accessibility $L_2$, crowding $L_3$, safety $L_4$, and sightseeing $L_5$. A Pareto-dominance check removes labels dominated by others at each stop.

\begin{figure}[t]
\centering
\includegraphics[width=1\textwidth]{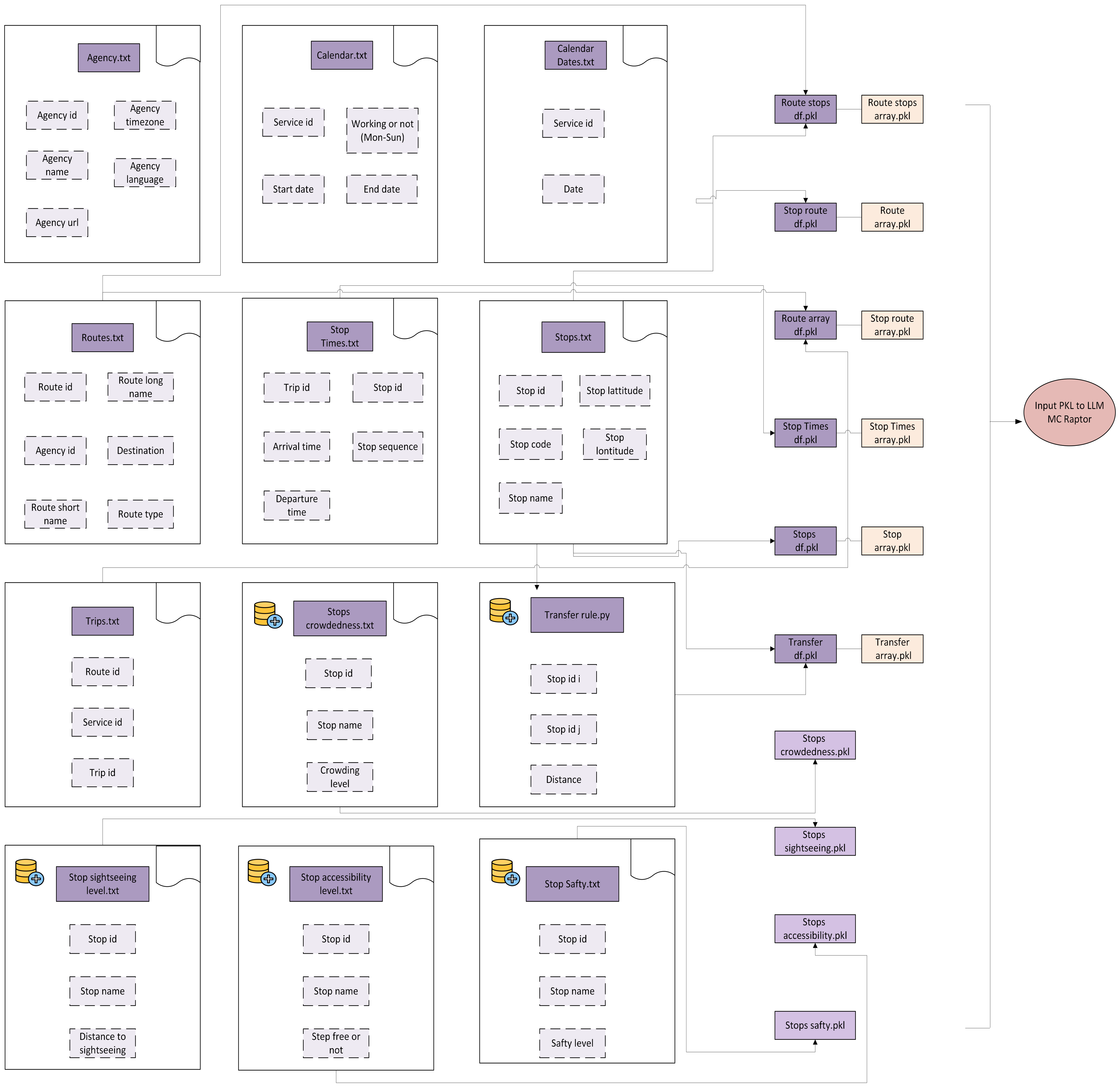}
\caption{Data Preparation for MC-RAPTOR}\label{mcraptor_data}
\end{figure}

The MC-RAPTOR algorithm is a timetable search algorithm that minimizes travel time and the number of transfers. To support the algorithm in considering additional and user-centric criteria, such as accessibility, crowding, safety, and sightseeing, this study integrates additional datasets as shown in Figure~\ref{mcraptor_data}. In particular, the accessibility level, crowding level, safety level, and sightseeing level are the extra criteria considered in this study. The selected active preferences determined in Node 3 directly weight the third and fourth criteria.

Crowding and safety levels at each station are simulated using a normal distribution and then normalized to the range $[0,1]$, where 0 represents the lowest level and 1 represents the highest level. Tourist-friendly level is based on whether the distance of each station to the London tourist attractions is within 100 meters. The London tourist locations are downloaded from the Open Trip Map by enabling the tourism option \citep{opentripmap2025}. The step-free data are collected from Transport for London (TfL) to assess accessibility of stops; all buses and bus stops are step-free, featuring the low-floor vehicles with ramps that deploy to make boarding easier. The step-free data are mainly collected for the underground and tram \citep{TfL2025}. The timetable data is based on the London city TfL GTFS data from June 2024 to December 2024.

\section{Experiments and Discussions}
\label{sec:experiments}

To validate the effectiveness of our proposed ChatPlanner, four complementary experiments are designed to evaluate different components and capabilities. Experiment 1 investigates and compares the feasibility of end-to-end route planning between LLM-based approaches and our ChatPlanner framework. Experiment 2 evaluates the LLM parsing component's ability to extract travel information and predict user preferences from natural language requests. Experiment 3 demonstrates how ChatPlanner enlarges the solution space by finding diverse routes that serve different user preferences beyond traditional optimization criteria. Experiment 4 assesses the viability of ChatPlanner for real-world deployment by measuring the latency and computational tractability of both the language processing and route search blocks.

\subsection{Experiment 1: Route Planning Feasibility by LLM with Tool}\label{sec:exp2}

This experiment aims to evaluate the end-to-end public transit routing performance of the available LLM equipped with public transit routing tools. Specifically, whether it can generate realistic, feasible, and complete transit routes when given the same journey planning queries as ChatPlanner.

GPT-5.0 (released in August 2025) is utilized and accessed via the OpenAI API in September 2025 \citep{singh2025openai}, a state-of-the-art closed-source model. For each test case, GPT-5.0 first parses the origin, destination, and requested arrival time from the natural language query, then calls a Google Maps API tool, which is implemented as a Python script that receives the parsed variables from LLM, queries the API, and returns the structured routing response to the LLM. If the Google Maps API does not return a valid result, GPT-5.0 is instructed to use its web search tool to query Moovit \citep{moovit2025} and Rome2Rio \citep{rome2rio2025} as fallback routing services. All three services contain the whole London GTFS data used by ChatPlanner.

\paragraph{Test Dataset}
45 basic journey planning query instances are constructed using the template "I need to get from \{start\} to \{end\} by \{time\}" with actual London locations and realistic arrival times. Each query contains information: start and end stops, and the requested arrival time.

For each case, the origin and destination routes were manually verified using Citymapper and Google Maps, with the requested arrival time specified accordingly. For routes that did not appear in either map solver but were returned by the tool-augmented LLM, further verification was conducted by searching the route number of each segment against the London TfL dataset to confirm its existence.

\subsubsection{Evaluation Metrics}

To assess practical feasibility, the generated route planning solutions are evaluated across four key performance indicators (KPIs). These KPIs determine whether a generated route is useful for real transit users. Table~\ref{tab:error-dimensions} provides the scope for each KPI.

(i) Station Identification (SI) captures inaccuracies in location specification. This includes wrong station names for origins, destinations, or transfers; vague descriptions where no explicit station name is provided; and ambiguous stop references that prevent a valid API call from being issued. Such issues prevent the system from retrieving route information and prevent users from identifying the correct physical locations to begin or continue their journeys. (ii) Timetable Accuracy (TA) captures inaccuracies in timing information for valid services. This includes wrong departure or arrival times and incorrect duration estimates. TA excludes cases where the route itself does not exist. Timetable inaccuracies mean users may miss the timed service, though the route itself may remain feasible via an alternative departure on the same route. (iii) Route Validity (RV) captures wrong or unsupported services that do not serve the specified journey, and missing transit legs that leave a route incomplete. RV excludes timing issues for otherwise valid services. (iv) Walking Completeness (WC) concerns the pedestrian portions of a route, including incorrect estimates of walking time in a segment. 

$$\text{KPI Rate} = \frac{\text{Number of solutions with specific KPI type}}{\text{Total number of solutions}}$$

The Feasibility Rate serves as our overall success metric, representing the proportion of solutions that can actually be executed by real users:

$$\text{Feasibility Rate} = \frac{\text{Number of feasible solutions}}{\text{Total number of solutions}}$$

A solution is considered feasible if a user can complete the trip from the origin to the destination.

\begin{table}[htbp]
\footnotesize
\centering
\caption{Scope and exclusion criteria for each error dimension.}
\label{tab:error-dimensions}
\setlength{\tabcolsep}{4pt}
\renewcommand{\arraystretch}{1.15}
\begin{tabularx}{\textwidth}{>{\RaggedRight\arraybackslash}p{3.2cm}
                              >{\RaggedRight\arraybackslash}X
                              >{\RaggedRight\arraybackslash}p{3.6cm}}
\toprule
\makecell[l]{\scriptsize\textbf{Dimension}} &
\makecell[l]{\scriptsize\textbf{Scope}} &
\makecell[l]{\scriptsize\textbf{Exclude}} \\
\midrule
Station Identification (SI)
  & Wrong name, vague description, ambiguous stop for OD pair that blocks calling for external tools
  & - \\
Timetable Accuracy (TA)
  & Wrong departure/arrival times, wrong duration
  & Route does not exist  \\
Route Validity (RV)
  & Any public transit portion being invalid or incomplete, e.g., wrong service, unsupported line, and missing transit leg
  & Timings of a valid service \\
Walking Completeness (WC)
  & Covers walking portion: wrong walking time estimates and missing walking segment
  & Missing public transport segment that makes a transfer impossible  \\
\bottomrule
\end{tabularx}
\end{table}

\subsubsection{Performance Discussion and Comparison}

Table~\ref{tab:error-analysis-detail} presents a case-by-case summary of the KPIs observed for the tool-augmented GPT-5.0. Despite having access to external routing APIs and web search, the overall feasibility rate is only 35\%. RV records the highest value among all KPIs, achieving 37\%.

\begin{table}[htbp]
\footnotesize
\centering
\caption{Analysis of solutions from the Tool-Augmented LLM across 45 test cases (Cases 0--29, continued Table~\ref{tab:error-analysis-detail-continued}). S indicates a feasible route.}
\label{tab:error-analysis-detail}
\setlength{\tabcolsep}{3pt}
\renewcommand{\arraystretch}{1.15}
\begin{tabularx}{\textwidth}{c c c c c c c >{\scriptsize\RaggedRight\arraybackslash}X}
\toprule
\makecell{\scriptsize\textbf{Case}} &
\makecell{\scriptsize\textbf{Option}} &
\makecell{\scriptsize\textbf{SI}} &
\makecell{\scriptsize\textbf{TA}} &
\makecell{\scriptsize\textbf{RV}} &
\makecell{\scriptsize\textbf{WC}} &
\makecell{\scriptsize\textbf{Feasible}} &
\makecell{\scriptsize\textbf{Evidence}} \\
\midrule
0  & 1 &            & TA &            & WC &   & Wrong time (17:21$\rightarrow$17:33); walking omitted on first leg \\
1  & 1 &            & TA & RV &            &   & Only first segment shown; wrong time (9:40$\rightarrow$10:32) \\
2  & 1 &            &            &            & WC &   & Walk 2 mins stated as 10 mins \\
3  & 1 &            &            &            & WC & S & Final walk 9 mins stated as 4 mins \\
\multirow{2}{*}{4}
   & 1 &            & TA &            &            & S & Told to leave 20 mins before earliest possible \\
   & 2 &            &            &            &            & S & \\
5  & 1 &            & TA & RV &            &   & 81--90 mins shown as 23 mins; bus fails destination \\
6  & 1 &            &            &            &            & S & Transfer very tight \\
7  & 1 &            &            & RV &            &   & Transfer should be Waterloo, not London Bridge \\
8  & 1 &            &            &            & WC &   & First walking leg is (15 mins) not mentioned \\
9  & 1 & SI &            & RV &            &   & API failed; transfer station wrong \\
10 & 1 &            &            & RV &            &   & Bus 206 does not support this journey \\
11 & 1 & SI &            & RV &            &   & API failed \\
12 & 1 &            &            &            & WC & S & Walk 6 mins stated as 1 min \\
13 & 1 & SI &            &            &            &   & API failed; station name ambiguous \\
14 & 1 & SI &            &            &            &   & API failed \\
\multirow{2}{*}{15}
   & 1 &            &            &            &            & S & \\
   & 2 &            &            & RV &            &   & Barbican wrong transfer \\
16 & 1 &            &            & RV &            &   & Specific route is missing \\
17 & 1 &            &     TA       &            &            & S & Start time should be 30 mins earlier \\
18 & 1 & SI           &            &            &            &  & Bus stop name not specified \\
19 & 1 & SI &            &            &            &   & API failed; station name ambiguous \\
20 & 1 & SI &            &            &            &   & API failed; station name ambiguous \\
21 & 1 & SI &            &            &            &   & API failed; station name ambiguous \\
22 & 1 &            &            & RV & WC &   & First segment missing; how to reach Marble Arch omitted \\
\multirow{2}{*}{23}
   & 1 &            &            &     RV       &            &  & Specific route not provided \\
   & 2 &            & TA & RV &            &   & Bus 115 does not serve this journey \\
24 & 1 &            & TA &            &            & S & Specific departure time not mentioned \\
25 & 1 & SI &            &            &            &   & Both start and end stations are ambiguous \\
\multirow{3}{*}{26}
   & 1 &            & TA &            &            & S & Need 50 mins but shown as 35--40 mins \\
   & 2 &            &            & RV &            &   & Bus 13 does not serve this journey \\
   & 3 &            &            &     RV       & WC &   & Walking missing on first leg; train and tube not specified \\
\multirow{2}{*}{27}
   & 1 &            &            &            &            & S & \\
   & 2 &            &            & RV &            &   & District line does not support this journey \\
28 & 1 &            &            &            &            & S & \\
\multirow{2}{*}{29}
   & 1 &            &            &            &            & S & \\
   & 2 &            &            & RV &            &   & Bus 73 does not serve the final leg \\
\bottomrule
\end{tabularx}
\end{table}

\begin{table}[htbp]
\footnotesize
\centering
\caption{Analysis of solutions from the Tool-Augmented LLM (continued, Cases 30--44). S indicates a feasible route.}
\label{tab:error-analysis-detail-continued}
\setlength{\tabcolsep}{3pt}
\renewcommand{\arraystretch}{1.15}
\begin{tabularx}{\textwidth}{c c c c c c c >{\scriptsize\RaggedRight\arraybackslash}X}
\toprule
\makecell{\scriptsize\textbf{Case}} &
\makecell{\scriptsize\textbf{Option}} &
\makecell{\scriptsize\textbf{SI}} &
\makecell{\scriptsize\textbf{TA}} &
\makecell{\scriptsize\textbf{RV}} &
\makecell{\scriptsize\textbf{WC}} &
\makecell{\scriptsize\textbf{Feasible}} &
\makecell{\scriptsize\textbf{Evidence}} \\
\midrule
30 & 1 &    &    & RV &    &   & Bus 428 does not serve this OD pair \\
31 & 1 &    &    &    &    & S & \\
32 & 1 &    &    & RV &    &   & Bus 410 does not serve this OD pair \\
33 & 1 &    & TA &    &    & S & Total time is 41 mins not 25 mins \\
34 & 1 &    &    &    & WC &   & Second segment requires 20 mins walk; walking omitted \\
35 & 1 &  SI  &    &    &    &   & Ambiguous origin; failed to call Google API \\
36 & 1 & SI &    &    &    &   & ``Dixon Way'' is ambiguous \\
37 & 1 &    &    &    &    & S & \\
38 & 1 &    &    &    &    & S & \\
39 & 1 & SI &    &    &    &   & Failed to identify start point \\
40 & 1 &    &    &    & WC & S & Missing second walking segment (16 mins walk) \\
41 & 1 &    &    &    &    & S & \\
42 & 1 &    &    & RV &    &   & Final bus segment is missing \\
43 & 1 & SI &    &    &    &   & Google API failed; station name ambiguous \\
44 & 1 &    &    & RV &    &   & Invalid solution \\
\midrule
\multicolumn{2}{l}{\scriptsize Total (45 cases, 52 options)} & 13 & 9 & 19 & 9 & 18 & \\
\multicolumn{2}{l}{\scriptsize Error / feasibility rate}     & 25\% & 17\% & 37\% & 17\% & 35\% & \\
\bottomrule
\end{tabularx}
\end{table}

\begin{table}[h]
\small
\centering
\caption{Summary comparison of GPT-5.0 with Tools and ChatPlanner across 45 test cases.}
\label{tab:summary-comparison}
\setlength{\tabcolsep}{3pt}
\renewcommand{\arraystretch}{1.15}
\begin{tabularx}{\textwidth}{L i c c c c c}
\toprule
\makecell{\scriptsize\textbf{Method}} &
\multicolumn{1}{c}{\makecell{\scriptsize\textbf{Solution}\\\scriptsize\textbf{count}}} &
\makecell{\scriptsize\textbf{SI}\\\scriptsize\textbf{rate}} &
\makecell{\scriptsize\textbf{TA}\\\scriptsize\textbf{rate}} &
\makecell{\scriptsize\textbf{RV}\\\scriptsize\textbf{rate}} &
\makecell{\scriptsize\textbf{WC}\\\scriptsize\textbf{rate}} &
\makecell{\scriptsize\textbf{Feasibility}\\\scriptsize\textbf{rate}} \\
\midrule
GPT-5.0 with Tools & 52 & 25\% & 17\% & 37\% & 17\% & 35\% \\
ChatPlanner        & 73 &  1\% &  0\% &  0\% &  0\% & 99\% \\
\bottomrule
\end{tabularx}
\end{table}

A primary cause of this low feasibility is the inherent instability of autonomous tool use. When querying the Google Maps API, the LLM can generate malformed requests to the API even when query codes and formats are provided. When forced to fall back on web search tools (e.g., Moovit or Rome2Rio), the model must parse complex, unstructured web page layouts. This extraction process is prone to information loss. If the LLM misses an intermediate step or misinterprets the retrieved text, it leads to an invalid route. These manifest as incorrect transfer stations (e.g., London Bridge instead of Waterloo in Case~7), bus route misassignments where real service numbers are applied to segments they do not operate (e.g., Bus 13 and 73 in Cases~26 and~29), and completely omitted transit legs (Cases~16 and~22).

SI further illustrates the challenge of using an LLM without a tightly integrated location database. The location names the LLM extracts from the user's prompt often do not exactly match the location strings expected by the external routing tools. For example, Cases~19-21 and 25 fail because the LLM submits ambiguous station names that the external APIs cannot resolve, returning zero results or routing the user to an incorrect location.

Furthermore, TA and WC occur when the model incorrectly transcribes and summarizes the retrieved data. For instance, in Case~5, the LLM extracts an incorrect travel duration (stating 23 minutes for an 80-minute journey), rendering the suggested departure time unusable. LLM could overlook walking details during summarization, resulting in omitted walking segments (Case~0) or incorrect walking time estimates (Case~12).

Table~\ref{tab:summary-comparison} compares these baseline results against the proposed ChatPlanner framework. ChatPlanner achieves a 99\% feasibility rate across 73 generated solutions, significantly outperforming the tool-augmented LLM. This performance gap is a direct result of ChatPlanner's modular architecture. Rather than relying on the LLM to execute tool calls or summarize path planning results, ChatPlanner strictly limits the LLM's role to natural language parsing and conversation handling. The extracted location names are handled and mapped to the same stops locations database used by the routing algorithm, which minimizes API mismatches and reduces station identification errors to 1\%. Once the locations are accurately resolved, the deterministic MC-RAPTOR algorithm computes the routes. Because the actual path planning is handled by an exact search algorithm rather than language generation, ChatPlanner is not affected by route validity, timetable accuracy, and walking completeness errors (0\% across all three dimensions).

\subsection{Experiment 2: User Request Parsing}
\label{sec:exp1}

This experiment evaluates the performance of the LLM component in Node 1 of our system. The LLM processes natural language requests and produces structured JSON output for three tasks: (1) extraction of stop names and request arrival time, (2) regression of user preference scores for four criteria, and (3) active preferences selection, involving multi-label classification and ranking of criteria based on the regression scores. Tasks (1) and (2) directly reflect the LLM component effectiveness for understanding and parsing the user request in Node 1, while task (3) reflects the effectiveness for guiding subsequent criteria selection in Node 3. Finally, to explicitly validate the value of the LLM's semantic understanding, LLM's rubric-consistent preference interpretation performance is compared against a traditional keyword matching system as the baseline.

\subsubsection{Experimental Setup}\label{sec:exp2_setup}


The test set comprises 120 instances collected through a user study. Participants were asked to provide transit journey requests, organized across five travel scenarios: an everyday journey, a special occasion, a journey planned to help another person, a journey under an unusual constraint, and a free-choice journey. Participants completed all ten requests before receiving any information about the four preferences (accessibility, crowdedness, safety, and sightseeing). After completing all requests, participants were introduced to the four preferences with a scoring rubric and scored example transit requests. Each preference is scored on a continuous scale from 0 to 1 for each of their ten requests. Participants also provided the origin, destination, and time request for each instance. From the 17 returned questionnaires, five were excluded (one blank, two missing preference labels, and two incomplete), yielding 120 labeled requests from 12 fully valid participants. These labeled requests were then processed into the User Study Test Set, where each instance consists of a natural language travel request as the instruction, and a structured label containing the origin, destination, time specification, and four preference scores as the output. In evaluation, the models are given each travel request and are expected to generate the corresponding output.

Three properties distinguish this user study test set from our training and RAG data. First, the requests capture the essences of public transit queries, including ambiguous expressions and mixed priorities, as they were composed freely without template constraints or domain-specific priming. Second, because participants wrote their requests without prior knowledge of the four preferences, the request text carries no systematic bias toward those preferences. Third, participants had no access to the eight personas and five travel contexts used for preference injection in constructing the fine-tuning and RAG datasets; the test set is therefore out-of-distribution (OOD) with respect to both, providing a more realistic assessment of generalization than evaluation based on in-distribution synthetic data, where inputs map cleanly to predefined personas or contexts.

To assess label reliability, three independent annotators scored all 120 instances across the four preferences using the same rubric and the same examples provided to the original participants. The intraclass correlation coefficient (ICC(A,1), two-way random, absolute agreement, single measures \citep{mcgraw1996forming}) was computed across three annotators and the original participant, following \citet{koo2016guideline}. The ICC values were 0.79 (95\% CI: [0.76, 0.82]) for accessibility, 0.75 ([0.71, 0.80]) for crowdedness, 0.77 ([0.74, 0.79]) for safety, and 0.81 ([0.78, 0.83]) for sightseeing, indicating moderate to good inter-annotator agreement \citep{artstein2008survey, koo2016guideline} and supporting the use of the original participant scores as ground truth in Experiment 2.

Two open-source LLMs are selected, including Llama3-8B-Instruct \citep{dubey2024llama} and Qwen2.5-7B-Instruct \citep{qwen2025qwen25technicalreport}. Open-source models differ from closed-source models in that they allow access to model weights and architecture, enabling fine-tuning for domain-specific tasks. 

\subsubsection{Experimental Design}
Four different model configurations are compared: (i) ChatPlanner (LoRA fine-tuned LLM with RAG), (ii) raw base LLM, (iii) base LLM with RAG, and (iv) LoRA fine-tuned LLM without RAG.

For each data instance in the test set, the journey request is input into the parsing component of each configuration, and the output JSON is compared with the ground truth label across three evaluation tasks: (1) exact matching, (2) preference score regression, and (3) active preferences selection.

\subsubsection{Task 1: Exact Matching}

Task 1 examines the model's ability to extract the exact origin, destination names, and the requested arrival time from the user's natural language input query. Exact string matching is evaluated for stop names and requested arrival time values.

For discrete slots (start stop, end stop, requested arrival time), exact match accuracy is used as the evaluation metric, following the information extraction practice established in \citep{rajpurkar2016squad}.

\textbf{Exact match accuracies} for stop names (\textbf{EM start} and \textbf{EM end}) and requested arrival time (\textbf{EM time}) extraction directly measure the model's ability to extract origin and destination locations and requested arrival time without any tolerance for variation. They are calculated as

$$\text{EM}_{\text{stop}} = \frac{1}{N} \sum_{i=1}^{N} \mathbbm{1}[\hat{s}_i = s_i], \qquad \text{EM}_{\text{time}} = \frac{1}{N} \sum_{i=1}^{N} \mathbbm{1}[\hat{t}_i = t_i]$$

\noindent where $N$ is the total number of test instances, $\hat{s}_i$ and $s_i$ are the predicted and true stop names, $\hat{t}_i$ and $t_i$ are the predicted and true time values for instance $i$, and $\mathbbm{1}[\cdot]$ is the indicator function that returns 1 if the condition is true and 0 otherwise. 

\textbf{EM all} is additionally defined as the joint exact match, where all three fields must be simultaneously correct:
$$\text{EM}_{\text{all}} = \frac{1}{N} \sum_{i=1}^{N} \mathbbm{1}[\hat{s}_i^{\text{start}} = s_i^{\text{start}} \wedge \hat{s}_i^{\text{end}} = s_i^{\text{end}} \wedge \hat{t}_i = t_i]$$
where $\wedge$ denotes logical AND.

A time tolerance accuracy is also included as an operational relaxation for practical transportation applications. \textbf{Time tolerance accuracy} allowing 5-minute deviation (\textbf{TimeTol@5mins}) provides a measure with tolerance, as small time deviations are sometimes acceptable in real-world scenarios. It is calculated as

$$\text{TimeTol@5mins} = \frac{1}{N} \sum_{i=1}^{N} \mathbbm{1}[|\hat{t}_i - t_i| \leq 5]$$

Table~\ref{tab:exact-match-stops-time-humantest} presents the exact matching metrics for stop name and requested arrival time extraction. These metrics measure the model's ability to parse discrete fields from natural language input.

\begin{table}[htbp]
\small
\centering
\caption{Exact matching performance for stop names and requested arrival time.}
\label{tab:exact-match-stops-time-humantest}
\setlength{\tabcolsep}{3pt}
\renewcommand{\arraystretch}{1.05}
\begin{tabularx}{\textwidth}{L d d d d d}
\toprule
\makecell{\scriptsize\textbf{Label}} &
\multicolumn{1}{c}{\makecell{\scriptsize\textbf{EM}\\\scriptsize\textbf{start}}} &
\multicolumn{1}{c}{\makecell{\scriptsize\textbf{EM}\\\scriptsize\textbf{end}}} &
\multicolumn{1}{c}{\makecell{\scriptsize\textbf{EM}\\\scriptsize\textbf{time}}} &
\multicolumn{1}{c}{\makecell{\scriptsize\textbf{Time}\\\scriptsize\textbf{Tol@5mins}}} &
\multicolumn{1}{c}{\makecell{\scriptsize\textbf{EM}\\\scriptsize\textbf{all}}}
\\
\midrule
Llama Base        & 0.867 & 0.733 & 0.642 & 0.675 & 0.475 \\
Qwen Base         & 0.875 & 0.750 & 0.800 & 0.850 & 0.608 \\
Llama Base + RAG  & 0.858 & 0.808 & 0.850 & 0.892 & 0.692 \\
Qwen Base + RAG   & 0.883 & 0.833 & 0.917 & 0.958 & 0.775 \\
Llama Fine-tuned          & 0.892 & 0.875 & 0.908 & 0.958 & 0.808 \\
Qwen Fine-tuned           & 0.900 & 0.892 & 0.900 & 0.950 & 0.825 \\
\textbf{Llama Fine-tuned + RAG}    & 1.000 & 0.992 & 0.992 & 1.000 & \textbf{0.983} \\
Qwen Fine-tuned + RAG     & 1.000 & 1.000 & 0.975 & 1.000 & 0.975 \\
\bottomrule
\end{tabularx}
\end{table}

The results show that the base LLMs struggle with exact matching on the User Study Test Set, achieving only 47.5-60.8\% overall accuracy. Adding RAG components improves performance to 69.2-77.5\%, but still leaves significant parsing errors. The exact parsing of stops and time is important, and any error at this stage will result in route planner failure. After fine-tuning with LoRA, accuracy improves to 80.8-82.5\%. Combining fine-tuning with RAG leads to the best performance at 97.5-98.3\%. These results demonstrate their complementary roles. Fine-tuning enforces the strict output structure, while RAG provides query-specific examples to resolve location-specific, imprecise, conversational, and implicit human expressions, making their combination necessary for more reliable exact matching. The fine-tuning corpus draws from both US-context (SGD) and UK-context (MultiWOZ) data, which introduces variation in location-specific terminology. However, the RAG component helps ground the stop name resolution directly in the target TfL corpus. As shown in Table~\ref{tab:exact-match-stops-time-humantest}, while the fine-tuned models alone perform well, adding the RAG component achieves near-perfect stop name extraction (up to 100.0\% EM start and end). This demonstrates that the retrieval mechanism can compensate for distributional mismatches in transport terminology (e.g., US vs. UK vocabulary) between the training and test domains.

\subsubsection{Task 2: Preference Score Regression}

Given the user query, the model predicts one numerical score between 0 and 1 for each of the four criteria: accessibility, crowding, safety, and sightseeing, which reflect user preferences in public transit route planning. Standard regression metrics are used for evaluation, along with agreement measures \citep{hyndman2006another}.

\textbf{Mean Absolute Error (MAE)} measures the average magnitude of prediction errors for preference scores, providing an intuitive measure of how far predictions deviate from actively expressed user preferences. $N$ is the total number of test instances, $\hat{y}_i$ and $y_i$ are the predicted and true preference scores for instance $i$. MAE is calculated as 

$$\text{MAE} = \frac{1}{N} \sum_{i=1}^{N} |\hat{y}_i - y_i|$$


\textbf{Root Mean Square Error (RMSE)} penalizes larger errors more heavily than MAE, indicating how well the model avoids substantial errors in estimating user preferences. It is calculated as

$$\text{RMSE} = \sqrt{\frac{1}{N} \sum_{i=1}^{N} (\hat{y}_i - y_i)^2}$$

\textbf{Symmetric Mean Absolute Percentage Error (sMAPE)} normalizes errors across different score scales, providing a scale-independent measure of relative prediction accuracy. It is calculated as

$$\text{sMAPE} = \frac{1}{N} \sum_{i=1}^{N} \frac{|\hat{y}_i - y_i|}{(|\hat{y}_i| + |y_i|)/2}$$

While error metrics (MAE, RMSE, sMAPE) measure prediction deviations, the \textbf{Concordance Correlation Coefficient (CCC)} evaluates whether predictions agree with true values in both their correlation and absolute scale \citep{lawrence1989concordance}. CCC is calculated as 

$$\text{CCC} = \frac{2\rho\sigma_{\hat{y}}\sigma_y}{\sigma_{\hat{y}}^2 + \sigma_y^2 + (\mu_{\hat{y}} - \mu_y)^2}$$ 

\noindent where $\rho$ is the Pearson correlation between predicted and true scores, $\sigma_{\hat{y}}$ and $\sigma_y$ denote their standard deviations, and $\mu_{\hat{y}}$ and $\mu_y$ denote their means. CCC ranges from -1 to 1, with 1 indicating perfect agreement, 0 no concordance, and negative values indicating systematic disagreement. CCC captures both relative ordering and magnitude matching, rewarding models that both rank correctly and match magnitudes. This is useful because a model might correctly rank preferences but systematically over/underestimate scores, which error metrics alone cannot fully reveal.

Table~\ref{tab:regression-pref-humantest} presents the regression metrics for predicting user preference scores averaged across the four criteria. 

\begin{table}[htbp]
\small
\centering
\caption{Regression performance for preference score prediction.}
\label{tab:regression-pref-humantest}
\setlength{\tabcolsep}{3pt}
\renewcommand{\arraystretch}{1.05}
\begin{tabularx}{\textwidth}{L d d d d}
\toprule
\makecell{\scriptsize\textbf{Label}} &
\multicolumn{1}{c}{\scriptsize\textbf{MAE}} &
\multicolumn{1}{c}{\scriptsize\textbf{RMSE}} &
\multicolumn{1}{c}{\scriptsize\textbf{sMAPE}} &
\multicolumn{1}{c}{\scriptsize\textbf{CCC}} \\
\midrule
Llama Base        & 0.406 & 0.480 & 0.507 & 0.076 \\
Qwen Base         & 0.411 & 0.501 & 0.475 & 0.060 \\
Llama Base + RAG  & 0.138 & 0.185 & 0.190 & 0.602 \\
Qwen Base + RAG   & 0.191 & 0.275 & 0.294 & 0.556 \\
Llama Fine-tuned          & 0.109 & 0.177 & 0.193 & 0.744 \\
Qwen Fine-tuned           & 0.180 & 0.252 & 0.232 & 0.653 \\
\textbf{Llama Fine-tuned + RAG}    & \textbf{0.098} & \textbf{0.142} & \textbf{0.158} & \textbf{0.779} \\
Qwen Fine-tuned + RAG     & 0.129 & 0.185 & 0.202 & 0.680 \\
\bottomrule
\end{tabularx}
\end{table}

The regression results demonstrate clear improvements from both RAG and fine-tuning (Table~\ref{tab:regression-pref-humantest}). Base LLMs show MAE values around 0.4, which represents substantial error given the $[0,1]$ score range. Adding either RAG or fine-tuning individually reduces MAE considerably, while the combination of our designed fine-tuning and RAG achieves the best performance across all four metrics. The best-performing configuration, Llama Fine-tuned + RAG, achieves an MAE of below 0.1 and a CCC near 0.8, indicating both accurate preference magnitude prediction and substantial correlation with human-annotated preference scores.

\paragraph{\textbf{Statistical Robustness of Tasks 1 and 2}}
To quantify the variability in LLM performance for the extraction (Task 1) and regression (Task 2) across different test samples, paired bootstrap resampling is applied, and confidence intervals are computed following standard practices \citep{tibshirani1993introduction, dror2018hitchhiker}. Since LLM performance can vary depending on which specific test examples are evaluated, a single test set may not fully represent the model's true capability. Bootstrap resampling addresses this limitation by drawing 1,000 resampled test sets of equal size (120 instances each) with replacement from the User Study Test Set, where some instances may appear multiple times, and others may be omitted in each resample. EM and MAE are recomputed on each resampled set, and the resulting 95\% confidence intervals (CIs), reported in Table~\ref{tab:metrics-bootstrap-em-mae-humantest}, indicate the range within which true performance likely falls.

\begin{table}[htbp]
\small
\centering
\caption{Bootstrapped EM and MAE (means and 95\% CIs).}
\label{tab:metrics-bootstrap-em-mae-humantest}
\setlength{\tabcolsep}{3pt}
\renewcommand{\arraystretch}{1.05}
\begin{tabularx}{\textwidth}{L d d d d d d}
\toprule
\makecell{\scriptsize\textbf{Label}} &
\multicolumn{1}{c}{\makecell{\scriptsize\textbf{EM all}\\\scriptsize\textbf{mean}}} &
\multicolumn{1}{c}{\makecell{\scriptsize\textbf{EM all}\\\scriptsize\textbf{CI low}}} &
\multicolumn{1}{c}{\makecell{\scriptsize\textbf{EM all}\\\scriptsize\textbf{CI high}}} &
\multicolumn{1}{c}{\makecell{\scriptsize\textbf{Score MAE}\\\scriptsize\textbf{mean}}} &
\multicolumn{1}{c}{\makecell{\scriptsize\textbf{MAE}\\\scriptsize\textbf{CI low}}} &
\multicolumn{1}{c}{\makecell{\scriptsize\textbf{MAE}\\\scriptsize\textbf{CI high}}}
\\
\midrule
Llama Base        & 0.475 & 0.408 & 0.558 & 0.406 & 0.383 & 0.428 \\
Qwen Base         & 0.608 & 0.567 & 0.649 & 0.411 & 0.389 & 0.432 \\
Llama Base + RAG  & 0.692 & 0.667 & 0.717 & 0.138 & 0.118 & 0.145 \\
Qwen Base + RAG   & 0.775 & 0.742 & 0.792 & 0.191 & 0.176 & 0.208 \\
Llama Fine-tuned          & 0.808 & 0.775 & 0.858 & 0.109 & 0.105 & 0.120 \\
Qwen Fine-tuned           & 0.825 & 0.775 & 0.867 & 0.180 & 0.164 & 0.195 \\
\textbf{Llama Fine-tuned + RAG}    & \textbf{0.983} & \textbf{0.925} & \textbf{1.000} & \textbf{0.098} & \textbf{0.090} & \textbf{0.109} \\
Qwen Fine-tuned + RAG     & 0.975 & 0.925 & 1.000 & 0.129 & 0.119 & 0.138 \\
\bottomrule
\end{tabularx}
\end{table}

The confidence intervals confirm two key findings. First, the performance gains from combining fine-tuning with RAG are statistically robust: both ChatPlanner configurations (LLM Fine-tuned + RAG) achieve EM CIs whose lower bounds remain above 0.92, with no overlap against any other configurations. This rules out the possibility that the observed improvements are accidentally achieved due to a few simple examples. Second, configurations with better performance also produce more stable results. The best configuration has a CI width of only 0.019 for MAE, whereas base LLMs without RAG exhibit CI widths more than two times larger, meaning their performance fluctuates more depending on which particular test instances are included. Together, these results provide statistical evidence that the combination of fine-tuning and RAG is not only more accurate but also more reliable across different test samples.

\subsubsection{Task 3: Active Preferences Selection}
In Node 3, the four predicted preference scores are used to select a subset of criteria to guide route planning. The selection process consists of (i) filtering out criteria with scores below 0.6, and (ii) selecting the top two highest-scoring criteria among the remaining criteria (or all remaining criteria if fewer than two remain after filtering). Task 3 is evaluated from two complementary perspectives: (i) as a multi-label classification problem and (ii) as a ranking preservation problem.

\paragraph{(I) As a multi-label classification problem} 
Each query can activate up to two criteria from accessibility, crowdedness, safety, and sightseeing, making this a multi-label classification problem of predicting which subset of criteria should be selected. Unlike standard classification, where each instance gets exactly one label, here each query can receive multiple labels simultaneously. For instance, a user might express preferences for both safety and accessibility. This experiment assesses whether the model selects the correct subset using (1) multi-label classification metrics \citep{madjarov2012extensive, zhang2013review, chicco2020advantages}; and (2) calibration metrics \citep{guo2017calibration,glenn1950verification}, to evaluate whether the model's confidence in its selections is well-calibrated. 

The multi-label classification metrics are defined as follows. 

\textbf{Micro-averaged F1 score (Micro-F1)} measures overall classification performance across all criteria by treating each criterion selection as an independent binary decision. It is calculated as

$$\text{Micro-F1} = \frac{2 \cdot \text{Micro-P} \cdot \text{Micro-R}}{\text{Micro-P} + \text{Micro-R}}$$

\noindent where Micro-P = $\frac{\sum_{j=1}^{L} TP_j}{\sum_{j=1}^{L} (TP_j + FP_j)}$ and Micro-R = $\frac{\sum_{j=1}^{L} TP_j}{\sum_{j=1}^{L} (TP_j + FN_j)}$, are micro-averaged precision and recall. $L$ is the number of criteria labels, and $TP_j$, $FP_j$, $FN_j$ are true positives, false positives, and false negatives for criterion $j$.

\textbf{Intersection over Union (Jaccard)} measures the similarity between predicted and true criteria sets, directly evaluating how well the model selects the correct combination of transportation criteria. $\hat{Y}_i$ and $Y_i$ are predicted and true label set. Jaccard is calculated as

$$\text{Jaccard} = \frac{1}{N} \sum_{i=1}^{N} \frac{|\hat{Y}_i \cap Y_i|}{|\hat{Y}_i \cup Y_i|}$$

\textbf{Exact match accuracy for label sets (Subset Accuracy)} measures the proportion of instances where the model selects exactly the correct set of criteria, providing the strictest evaluation of criteria selection performance. $\hat{Y}_i$ and $Y_i$ are the set of predicted and true criteria for instance $i$, and $\mathbbm{1}[\cdot]$ is the indicator function. Subset Accuracy is calculated as

$$\text{Subset Accuracy} = \frac{1}{N} \sum_{i=1}^{N} \mathbbm{1}[\hat{Y}_i = Y_i]$$

\textbf{Area Under the ROC Curve (AUC)} evaluates the model's ability to distinguish between criteria that should and should not be selected across different decision thresholds. It is calculated using the trapezoidal rule for each label and then averaged.



Calibration metrics are defined below. 

\textbf{Expected Calibration Error (ECE)} measures the difference between predicted confidence and actual accuracy, with lower values indicating better calibration \citep{guo2017calibration}. It is calculated as

$$\text{ECE} = \sum_{m=1}^{M} \frac{|B_m|}{N} |\text{acc}(B_m) - \text{conf}(B_m)|$$

\noindent where $M$ is the number of confidence bins, $B_m$ is the set of samples in bin $m$, $|B_m|$ is the number of samples in bin $m$, $N$ is the total number of samples, $\text{acc}(B_m)$ and $\text{conf}(B_m)$ are the accuracy and average confidence of samples in bin $m$.

\textbf{Brier Score} measures both calibration and sharpness of probability predictions, with lower values indicating better probabilistic predictions \citep{glenn1950verification}. $p_i$ and $y_i$ are the predicted probability and actual binary outcome (0 or 1) for sample $i$. The Brier score is calculated as

$$\text{Brier} = \frac{1}{N} \sum_{i=1}^{N} (p_i - y_i)^2$$


Table~\ref{tab:metrics-decisions-calib-humantest} presents performance measured by the multi-label classification metrics and calibration metrics for Task 3.

\begin{table}[htpb]
\small
\centering
\caption{Multi-label classification performance for criteria selection.}
\label{tab:metrics-decisions-calib-humantest}
\setlength{\tabcolsep}{3pt}
\renewcommand{\arraystretch}{1.05}
\begin{tabularx}{\textwidth}{L d d d d d d}
\toprule
\makecell{\scriptsize\textbf{Label}} &
\multicolumn{1}{c}{\scriptsize\textbf{micro F1}} &
\multicolumn{1}{c}{\scriptsize\textbf{Jaccard}} &
\multicolumn{1}{c}{\makecell{\scriptsize\textbf{Subset}\\\scriptsize\textbf{Accuracy}}} &
\multicolumn{1}{c}{\scriptsize\textbf{AUC}} &
\multicolumn{1}{c}{\scriptsize\textbf{ECE}} &
\multicolumn{1}{c}{\scriptsize\textbf{Brier}} \\
\midrule
Llama Base        & 0.516 & 0.403 & 0.142 & 0.563 & 0.535 & 0.519 \\
Qwen Base         & 0.483 & 0.372 & 0.125 & 0.510 & 0.547 & 0.558 \\
Llama Base + RAG  & 0.726 & 0.669 & 0.550 & 0.769 & 0.272 & 0.278 \\
Qwen Base + RAG   & 0.650 & 0.569 & 0.450 & 0.743 & 0.328 & 0.303 \\
Llama Fine-tuned          & 0.770 & 0.729 & 0.617 & 0.813 & 0.231 & 0.220 \\
Qwen Fine-tuned           & 0.668 & 0.601 & 0.558 & 0.767 & 0.275 & 0.275 \\
\textbf{Llama Fine-tuned + RAG}    & \textbf{0.781} & \textbf{0.746} & \textbf{0.675} & \textbf{0.848} & \textbf{0.217} & \textbf{0.170} \\
Qwen Fine-tuned + RAG     & 0.732 & 0.682 & 0.608 & 0.795 & 0.231 & 0.215 \\
\bottomrule
\end{tabularx}
\end{table}

The multi-label classification metrics reveal a direct link between regression accuracy in Task 2 and criterion selection quality. Configurations that better predict preference scores also more accurately select which criteria to activate. Base LLMs perform at a near-chance level, with subset accuracy below 0.15, meaning nearly every request would trigger route planning with incorrect criteria. RAG and fine-tuning each improve performance substantially, but even with either component alone, roughly half of requests still select the wrong criteria combination. The combined approach achieves the strongest results, with a subset accuracy of 0.675 from the best configuration. While not perfect, this is acceptable for the initial planning stage, as only about one-third of requests would require adjustment through follow-up interaction, either re-ranking solutions or rerunning the route planning algorithm.

Calibration metrics assess whether high-confidence predictions are indeed more likely to be correct. Base LLMs produce confidence estimates that are essentially uninformative, with ECE values above 0.5. Both RAG and fine-tuning substantially improve calibration, with fine-tuning being more effective of the two. The combined approach again achieves the best results, producing the lowest ECE and Brier scores across all configurations. This indicates that fine-tuning and retrieval are complementary. Fine-tuning aligns the model's internal confidence with actual correctness, while RAG provides grounding examples that further sharpen prediction reliability.

\paragraph{(II) As a ranking preservation problem} 
Beyond selecting the correct criteria, the model should also preserve the relative ordering of user preferences. For example, if a user expresses a stronger preference for safety (scoring 0.8) than crowdedness (scoring 0.5), the model should maintain this ranking even if both criteria are eventually filtered out or selected. Ranking quality is measured using three rank correlation coefficients, as shown below. All rank correlation metrics range from $-1$ to $1$, with higher values indicating better preservation of preference ordering.

\textbf{Spearman's rank correlation coefficient (Spearman $\rho$)} is a nonparametric metric that measures monotonic relationships between the predicted and true score rankings, robust to score rescaling \citep{spearman1961proof}. It is calculated as

$$\rho = 1 - \frac{6\sum d_i^2}{n(n^2-1)}$$

\noindent where $n$ is the number of criteria, and $d_i$ is the difference between the ranks of criterion $i$ in the predicted and true scores.

\textbf{Kendall's rank correlation coefficient (Kendall $\tau$)} provides an alternative measure of rank correlation that emphasizes pairwise rank agreements \citep{kendall1938new}. It is calculated as

$$\tau = \frac{n_c - n_d}{\frac{1}{2}n(n-1)}$$

where $n_c$ and $n_d$ are the number of concordant pairs (where both the predicted and true rankings agree on relative order) and discordant pairs.

\textbf{Pearson correlation coefficient (Pearson $r$)} serves as a reference for linear relationships between predicted and true scores, though it is less robust to monotonic transformations than the nonparametric alternatives \citep{lee1988thirteen}. It is calculated as

$$r = \frac{\sum (x_i - \bar{x})(y_i - \bar{y})}{\sqrt{\sum (x_i - \bar{x})^2 \sum (y_i - \bar{y})^2}}$$

To evaluate whether models preserve the relative ordering of user preference scores across criteria, rank correlation is examined through $\rho$, $\tau$, and $r$. These metrics are presented in Table~\ref{tab:calib-rank-humantest}.

\begin{table}[h]
\small
\centering
\caption{Ranking performance for criteria selection.}
\label{tab:calib-rank-humantest}
\setlength{\tabcolsep}{3pt}
\renewcommand{\arraystretch}{1.05}
\begin{tabularx}{\textwidth}{L d d d}
\toprule
\makecell{\scriptsize\textbf{Label}} &
\multicolumn{1}{c}{\scriptsize\textbf{Spearman}} &
\multicolumn{1}{c}{\scriptsize\textbf{Kendall}} &
\multicolumn{1}{c}{\scriptsize\textbf{Pearson}} \\
\midrule
Llama Base        & 0.112 & 0.100 & 0.076 \\
Qwen Base         & 0.161 & 0.138 & 0.130 \\
Llama Base + RAG  & 0.563 & 0.529 & 0.566 \\
Qwen Base + RAG   & 0.556 & 0.518 & 0.546 \\
Llama Fine-tuned          & 0.703 & 0.687 & 0.701 \\
Qwen Fine-tuned           & 0.625 & 0.602 & 0.646 \\
\textbf{Llama Fine-tuned + RAG}    & \textbf{0.717} & \textbf{0.691} & \textbf{0.738} \\
Qwen Fine-tuned + RAG     & 0.630 & 0.604 & 0.656 \\
\bottomrule
\end{tabularx}
\end{table}

Base LLMs show near-zero correlations, meaning they largely fail to distinguish which criteria a user considers more important. This is a critical limitation, since Task 3 depends on correctly prioritizing preferences. Both RAG and fine-tuning substantially improve ranking quality. The combined approach again achieves the strongest performance, with all three correlation measures around 0.7. In practical terms, this means that when a user emphasizes safety over crowdedness, for example, the model consistently reflects this ordering in its predicted scores. The results suggest that retrieval provides relevant preference context from similar past queries, while fine-tuning teaches the model to assign and order criteria weights appropriately.

In summary, the combination of fine-tuning and RAG consistently achieves the best performance across all three tasks, with bootstrap analysis confirming these gains are statistically robust. The two components serve complementary roles. Fine-tuning enforces structured output formatting and teaches the model to map natural language to preference scores, while RAG grounds predictions in query-specific examples that resolve ambiguous or location-specific expressions.

\subsubsection{Comparison Against a Keyword Matching Baseline}\label{sec:keyword-baseline}

Rule-based frameworks, such as bag-of-words and keyword matching, have long served as standard baselines in transportation research for extracting user preferences and network features from unstructured natural language data \citep{ali2019fuzzy}. To directly assess whether the semantic interpretation capacity of our ChatPlanner provides measurable gains over such approaches, a keyword matching baseline is developed, and both systems are evaluated on the same binary preference detection task.

\paragraph{Keyword Matching System Design}
The keyword matching system is designed as a binary relevance classifier reliant on a hand-crafted, domain-specific dictionary.

(i) \textit{Accessibility} (25 keywords, e.g., ``wheelchair'', ``step-free'', ``elevator''); 

(ii) \textit{Safety} (31 keywords, e.g., ``safe'', ``well-lit'', ``crime''); 

(iii) \textit{Crowdedness} (31 keywords, e.g., ``crowded'', ``rush hour'', ``personal space''); and 

(iv) \textit{Sightseeing} (39 keywords, e.g., ``scenic'', ``landmarks'', ``architecture''). 

The complete lists of curated keywords for each preference are provided in ~\ref{apx:keyword-lists}.

The classifier employs a token-level matching approach to prevent sub-word matching errors inherent to basic regular expressions \citep{zhang2018deep}. Utilizing the Natural Language Toolkit (NLTK) \citep{bird2009natural}, the user's journey request is first lowercased and segmented into discrete word tokens, from which contiguous $n$-grams are additionally generated to enable
matching of multi-word keywords (e.g., ``rush hour'', ``step-free''). If the intersection between the resulting tokens and $n$-grams and a
predefined keyword list is non-empty, that corresponding preference is activated (assigned a value of $1$); otherwise, it remains inactive (assigned a value of $0$).

\paragraph{Experimental Setup and Evaluation Metrics}
To evaluate ChatPlanner against the keyword matching system in the binary preference detection task, the User Study Test Set described in Section~\ref{sec:exp2_setup} is adapted by binarizing the human-labeled continuous ground-truth scores. A preference is considered active if its score is $\geq 0.6$, and inactive otherwise. To ensure a fair comparison, $0.6$ is chosen as the same threshold to convert the continuous preference scores generated by the LLM in ChatPlanner into binary active and inactive labels.

Performance is evaluated using three standard metrics:
\begin{itemize}
    \item \textbf{Exact Match Accuracy}: The proportion of user requests where the model correctly predicts the binary activation states for all four criteria simultaneously.
    \item \textbf{F1-Score}: The harmonic mean of precision and recall, providing a robust evaluation metric when positive and negative samples are highly imbalanced (e.g., when a specific travel constraint is rarely requested). It is mathematically formulated as:
    $$F_1 = \frac{2TP}{2TP + FP + FN}$$
    where $TP$, $FP$, and $FN$ denote the number of true positives, false positives, and false negatives, respectively. We report both the overall F1-score to capture global performance, and the per-dimension F1-scores to evaluate robustness across different preference types.
\end{itemize}


\paragraph{Results and Discussion}
Table~\ref{tab:keyword-llm-comparison} presents the comparative results of the keyword matching baseline and our fine-tuned LLMs equipped with RAG (Llama Fine-tuned + RAG and Qwen Fine-tuned + RAG). The table also reports Support, which is the number of active instances per criterion out of the 120 instances in the User Study Test Set.

\begin{table}[htbp]
\small
\centering
\caption{Performance comparison between the Keyword Matching Baseline and ChatPlanner on binary preference detection.}
\label{tab:keyword-llm-comparison}
\setlength{\tabcolsep}{4pt}
\renewcommand{\arraystretch}{1.15}
\begin{tabularx}{\textwidth}{L c c c c c c}
\toprule
\multirow{2}{*}{\makecell{\scriptsize\textbf{Model / Metric}}} & 
\multicolumn{2}{c}{\scriptsize\textbf{Overall Performance}} & 
\multicolumn{4}{c}{\scriptsize\textbf{Per-Criterion F1-Scores}} \\
\cmidrule(lr){2-3} \cmidrule(lr){4-7}
& \makecell{\scriptsize\textbf{Exact Match}\\\scriptsize\textbf{Accuracy}} & \makecell{\scriptsize\textbf{Overall}\\\scriptsize\textbf{F1-Score}} & \scriptsize\textbf{Accessibility} & \scriptsize\textbf{Crowdedness} & \scriptsize\textbf{Safety} & \scriptsize\textbf{Sightseeing} \\
\midrule
Keyword Matching System Baseline & 0.242 & 0.560 & 0.590 & 0.619 & 0.460 & 0.571 \\
Qwen Fine-tuned + RAG & 0.650 & 0.833 & 0.800 & 0.833 & 0.848 & 0.889 \\
Llama Fine-tuned + RAG & 0.725 & 0.884 & 0.835 & 0.902 & 0.899 & 0.957 \\
\midrule
\textit{Support (Active Instances)} & \textit{-} & \textit{-} & \textit{41} & \textit{54} & \textit{63} & \textit{24} \\
\bottomrule
\end{tabularx}
\end{table}

The results show that keyword matching fails to identify the complete set of user preferences in the majority of cases, while ChatPlanner achieves substantially higher performance across both overall and per-criterion metrics. The performance gap is most significant for criteria that users tend to express implicitly rather than through explicit keywords. For example, a request such as "I am traveling with my grandmother" implies safety needs, and "I want a beautiful walk" implies sightseeing interest, but neither contains vocabulary that a keyword system would match, while the fine-tuned LLM with RAG can correctly identify these preferences. The per-criterion F1-scores confirm this pattern. Criteria where implicit expression could be more common (i.e., Safety and Sightseeing) show the largest gains over the keyword baseline. These findings demonstrate that the core value of integrating an LLM into the planning pipeline lies in its ability to interpret varied and indirect human expressions of travel preferences, ensuring the downstream routing solver optimizes for the correct objectives.

\subsubsection{Route-Level Preference Validation}\label{sec:route-pref-study}

Experiment~2 shows that ChatPlanner interprets \emph{stated} preferences accurately. A further
question is whether the routes it produces from those preferences are the routes users actually
prefer. To answer this, we conducted a validation test as a follow-up to Experiment~2. We pooled the solutions generated by ChatPlanner and the baseline MC-RAPTOR and asked the same participants to mark their top preferred route(s) for each journey. In total, 78~journeys received completed route selections from 12~participants, referred to as the \emph{User Route Choice Set}. As shown in Table~\ref{tab:route-pref-by-category}, the preferred route was
\emph{found only by ChatPlanner} in 68\% of journeys (53/78); in the remaining 32\% (25/78), the preferred route was produced by both ChatPlanner and the baseline. In 17 of these journeys, the two methods generated an identical route, shown in the \emph{shared solution} column, and in the other 8~journeys the participant selected one route from each method, which we label a \emph{shared tie}. This test confirms that ChatPlanner provides additional solutions that align with travelers' preferences.

\begin{table}[htbp]
\centering
\small
\caption{Route preference outcome of validation test}
\label{tab:route-pref-by-category}
\begin{tabular}{lcccc}
\toprule
Preference & $n$ & ChatPlanner-only & Shared solution & Shared Tie \\
\midrule
Overall & 78 & 53 (68\%) & 17 (22\%) & 8 (10\%) \\
\midrule
Accessibility & 26 & 20 (77\%) & 4 (15\%) & 2 (8\%) \\
Crowdedness & 42 & 28 (67\%) & 9 (21\%) & 5 (12\%) \\
Safety & 40 & 26 (65\%) & 9 (22\%) & 5 (12\%) \\
Sightseeing & 15 & 11 (73\%) & 4 (27\%) & 0 (0\%) \\
\bottomrule
\end{tabular}
\end{table}

\subsection{Experiment 3: ChatPlanner Enlarges Solution Space for User Preference}

ChatPlanner is compared against the baseline MC-RAPTOR algorithm, which only considers minimizing travel time and the number of transfers in journey planning. Case A focuses on mobility-impaired users requiring accessibility considerations. Case B examines commuters seeking less crowded routes. Case C explores tourist preferences for routes with sightseeing opportunities.

\begin{table}[htbp]
\small
\centering
\caption{Example user inquiries used for evaluation.}
\label{tab:user-inquiries}
\setlength{\tabcolsep}{3pt}
\renewcommand{\arraystretch}{1.25}
\begin{tabularx}{\textwidth}{W L}
\toprule
\makecell{\scriptsize\textbf{User Case}} &
\makecell{\scriptsize\textbf{User query}} \\
\midrule
Case A: Mobility-impaired people &
I'm traveling with a wheelchair and need elevator access. I want to plan my journey from Farringdon to Bow Road by 12:00. \\
Case B: Crowding-averse commuter &
As a daily commuter, I need spacious routes. I need to get from Charing Cross Underground Station to Euston Station by 15:45. \\
Case C: Tourist &
Getting around London: Sloane Square to Tate Britain by 10:00. I prefer routes that pass interesting architecture. \\
\bottomrule
\end{tabularx}
\end{table}

\begin{figure}[htpb]
\centering
\includegraphics[width=0.8\textwidth]{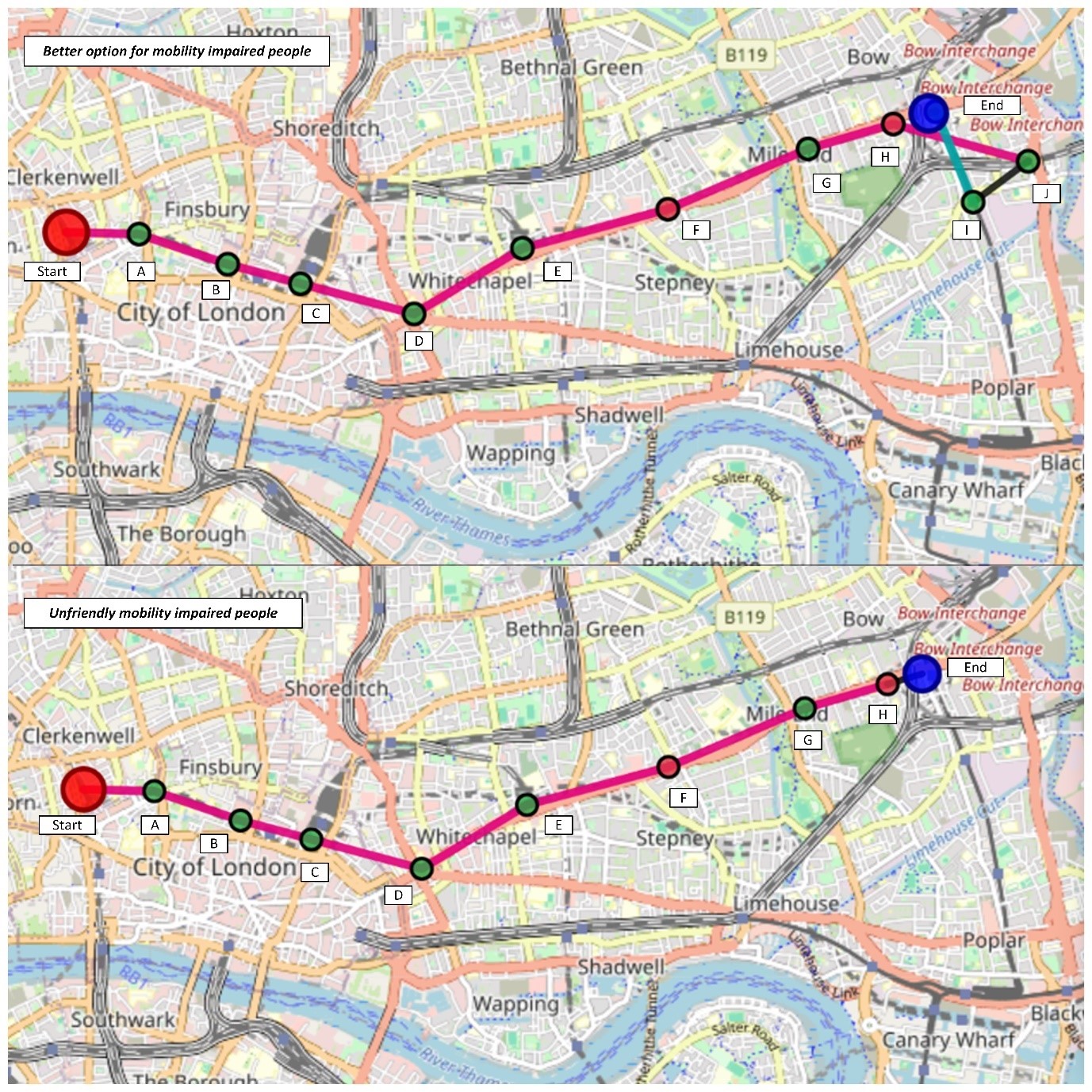}
\caption{Case A: Route Planning with Accessibility Need}\label{fig:case1}
\end{figure}

\paragraph{Case A: Accessibility for Mobility-Impaired Users}
For Case A, the journey is from Farringdon to Bow Road with a requested arrival time of 12:00. Figure~\ref{fig:case1} shows the routes. In the lower half figure, the baseline MC-RAPTOR algorithm finds a direct path that goes through fewer stations, start-A-B-C-D-E-F-G-H-End, but requires the user to exit at station H, which lacks step-free access. ChatPlanner (upper half) identifies an alternative solution with higher accessibility scores that suggests transferring at station I to reach another station J that both have step-free access. This route also ensures direct exiting at the end stop. Traditional optimization methods would reject this option because it requires more time and transfers. However, this alternative route provides a step-free option for users with limited mobility.

\begin{figure}[htpb]
    \centering 
    
    \begin{subfigure}{0.495\textwidth}
        \centering
        \includegraphics[width=\textwidth]{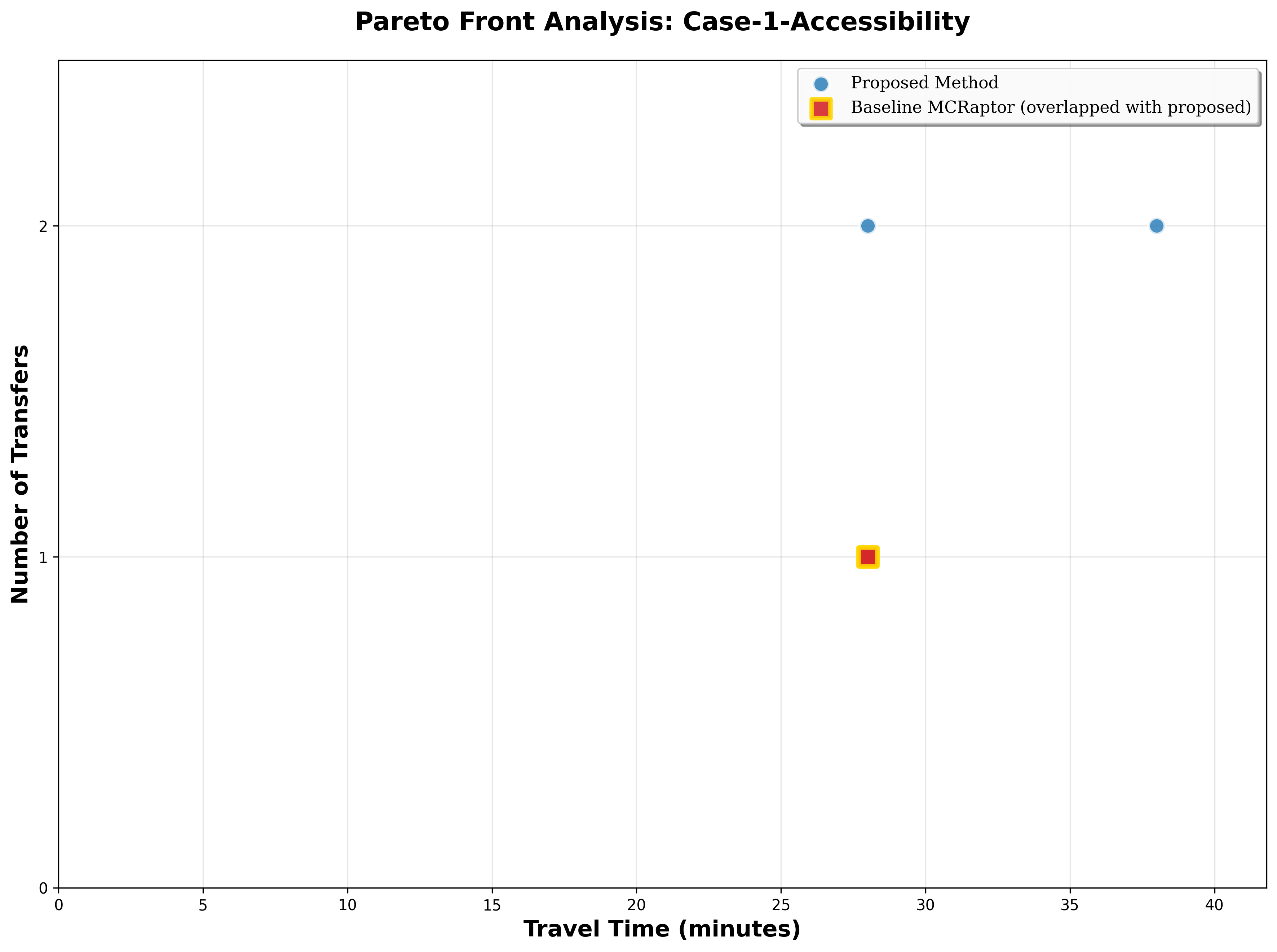}
        \caption{Case A: Pareto front comparison between ChatPlanner and MC-RAPTOR baseline in travel time vs number of transfers space.}
        \label{fig:case1_pareto_baseline}
    \end{subfigure}
    \hfill 
    \begin{subfigure}{0.495\textwidth}
        \centering
        \includegraphics[width=\textwidth]{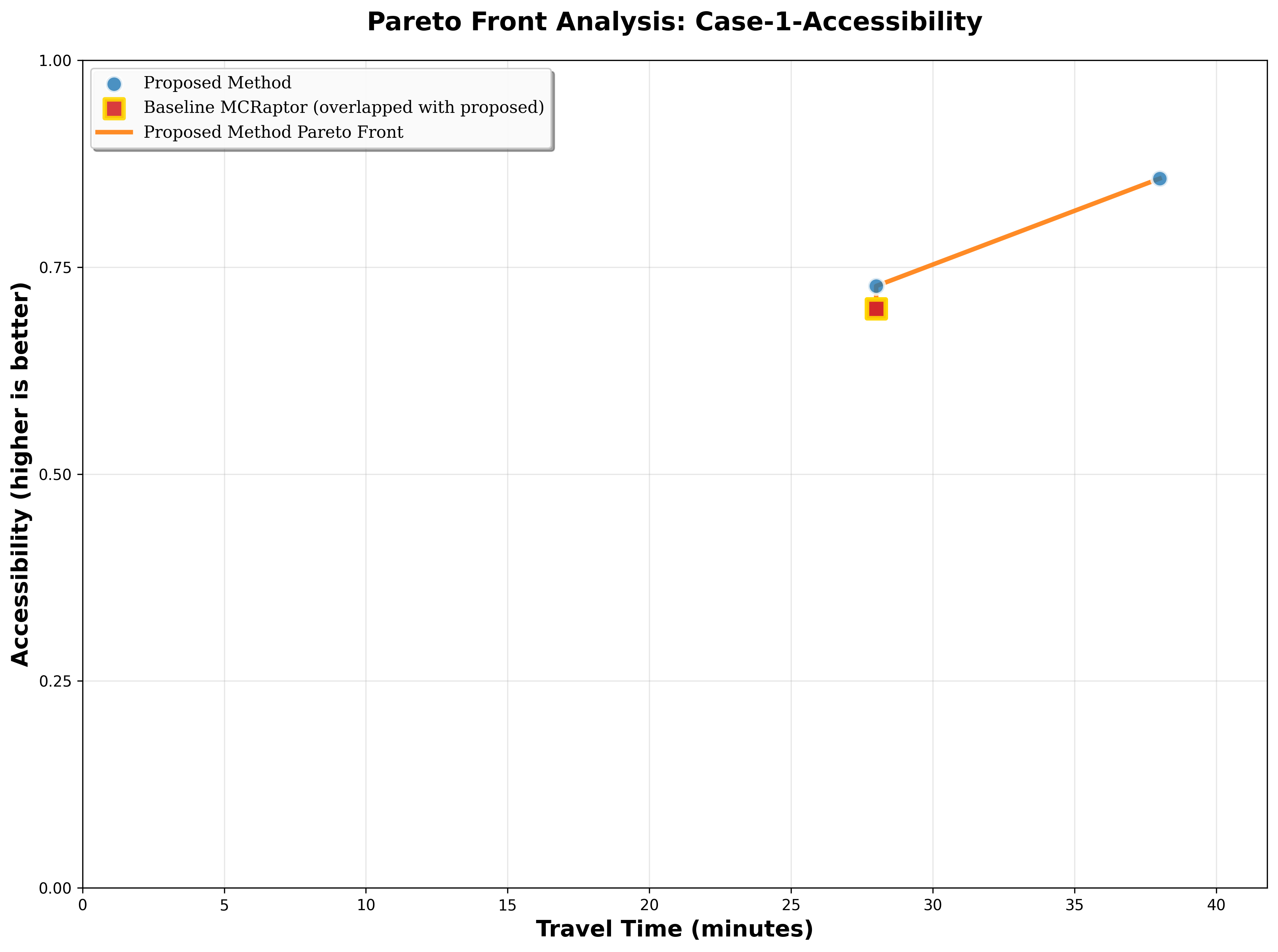}
        \caption{Case A: Pareto front in travel time vs accessibility, showing how ChatPlanner enlarges the solution space along accessibility.}
        \label{fig:case1_pareto_accessibility}
    \end{subfigure}
    
    \caption{Case A: Pareto front comparisons for ChatPlanner.}
    \label{fig:case1_pareto_combined} 
\end{figure}

\begin{figure}[htpb]
    \centering 
    
    \begin{subfigure}{0.495\textwidth}
        \centering
        \includegraphics[width=\textwidth]{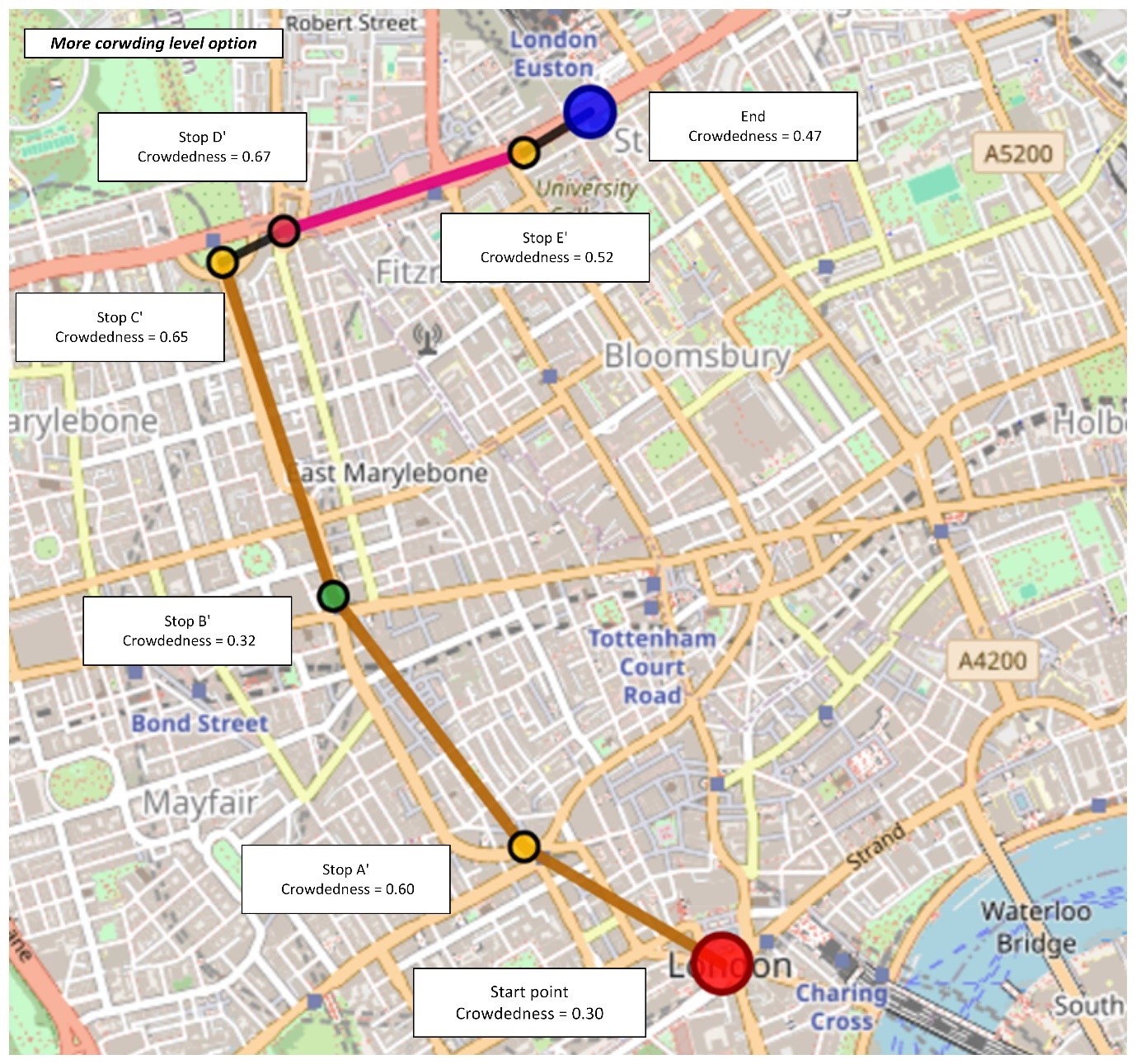}
        \caption{Case B: Route Planning with Crowding Need, } 
        \label{fig:case2_sol1}
    \end{subfigure}
    \hfill 
    \begin{subfigure}{0.495\textwidth}
        \centering
        \includegraphics[width=\textwidth]{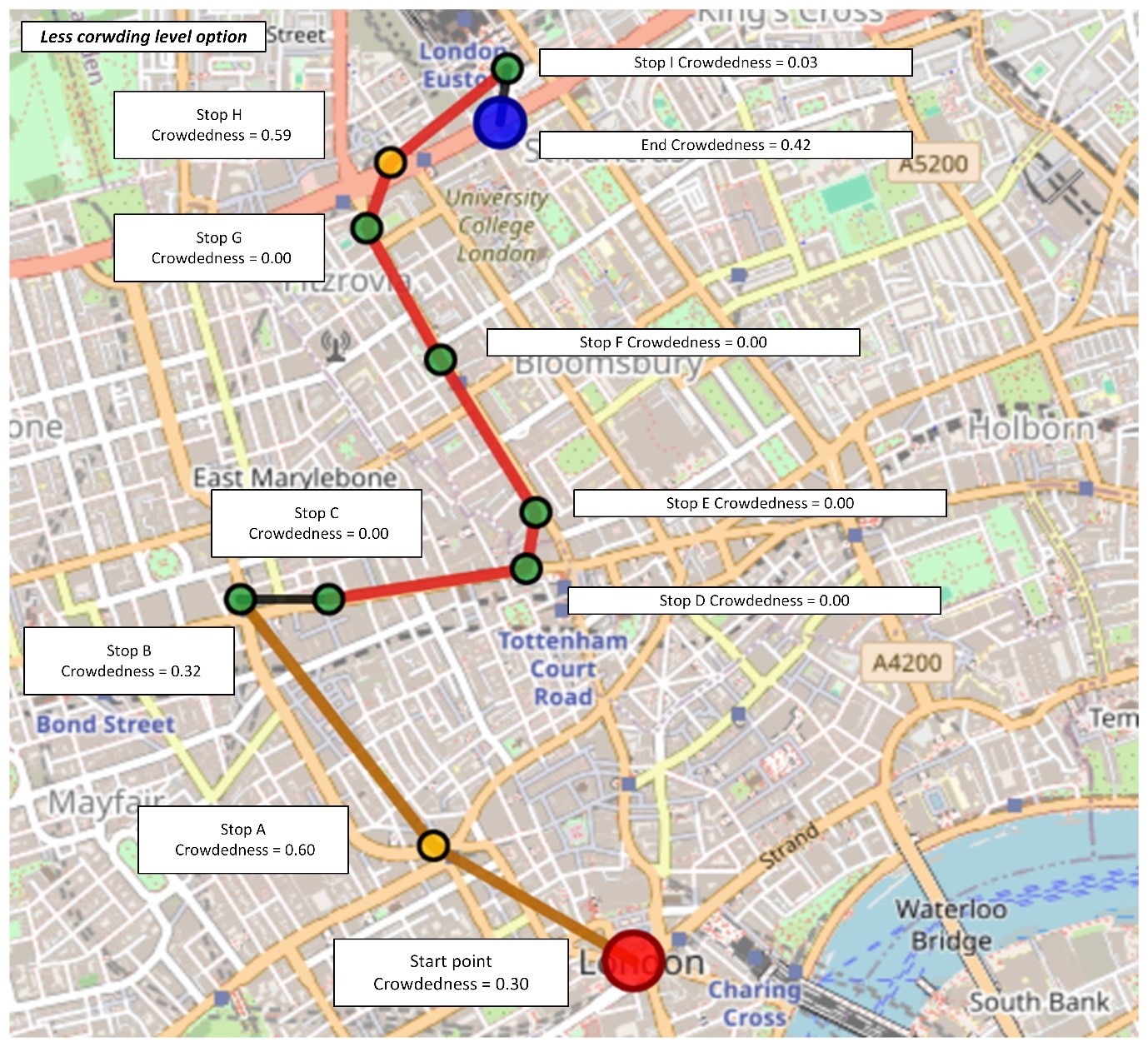}
        \caption{Case B: Route Planning with Crowding Need}
        \label{fig:case2_sol2}
    \end{subfigure}
    
    \caption{Case B: Route Planning with Crowding Need (two solutions).}
    \label{fig:case2_sols_combined} 
\end{figure}

\paragraph{Case B: Crowding Avoidance for Commuters}
In Case B, the user needs to travel from Charing Cross Underground Station to Euston Station by 15:45 while avoiding crowded areas. Figures~\ref{fig:case2_sol1} and \ref{fig:case2_sol2} compare two route options. Route 1 shows higher simulated crowding levels at most stops, while Route 2 maintains lower crowding throughout the journey. Although Route 2 requires more stops and takes longer travel time, it avoids stops with higher simulated crowding levels, providing a more comfortable travel experience for daily commuters who prioritize space over speed.

\begin{figure}[htpb]
    \centering 
    
    \begin{subfigure}{0.495\textwidth}
        \centering
        \includegraphics[width=\textwidth]{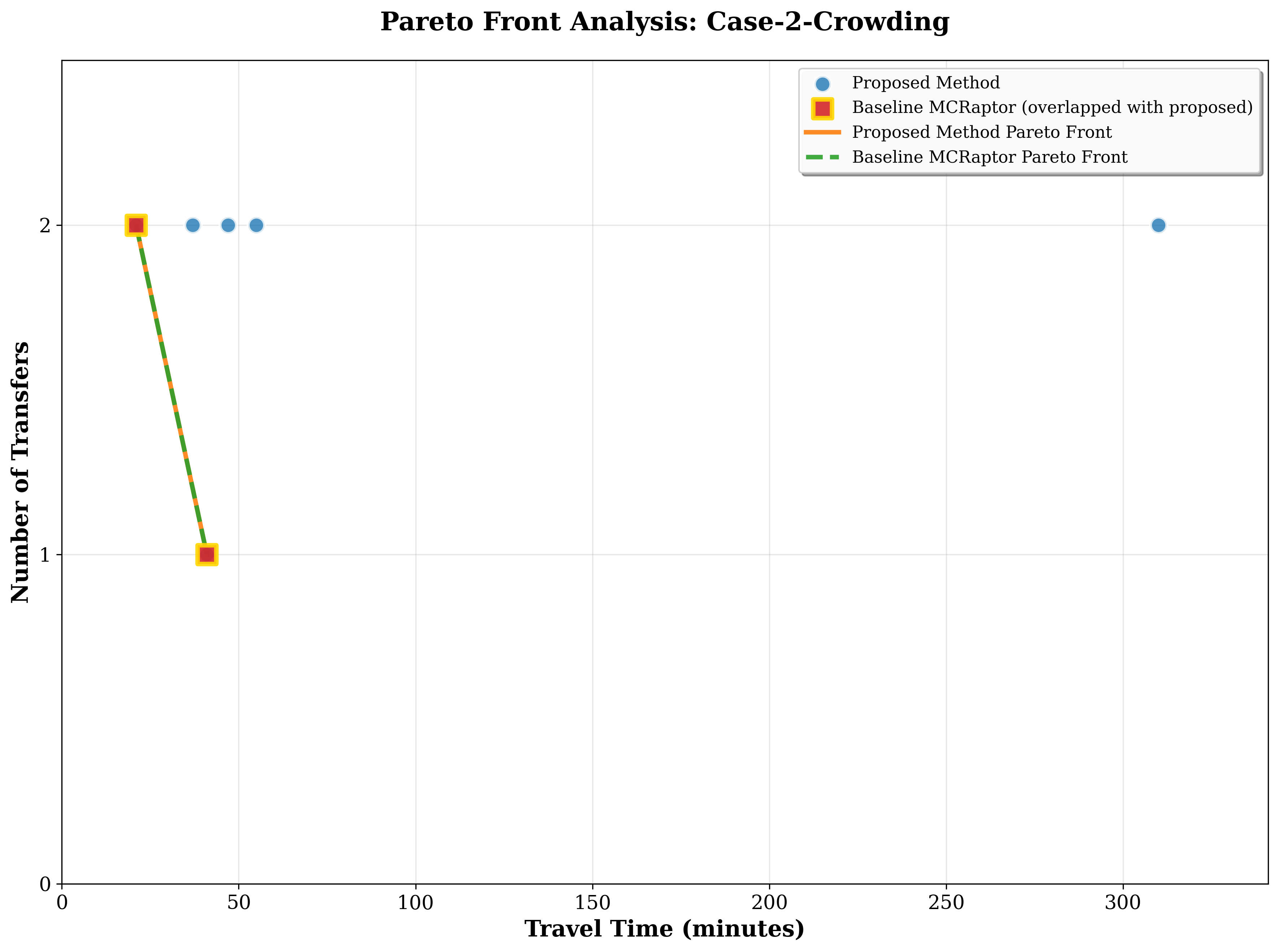}
        \caption{Case B: Pareto front comparison between ChatPlanner and MC-RAPTOR baseline in travel time vs number of transfers space.}
        \label{fig:case2_pareto_baseline}
    \end{subfigure}
    \hfill 
    \begin{subfigure}{0.495\textwidth}
        \centering
        \includegraphics[width=\textwidth]{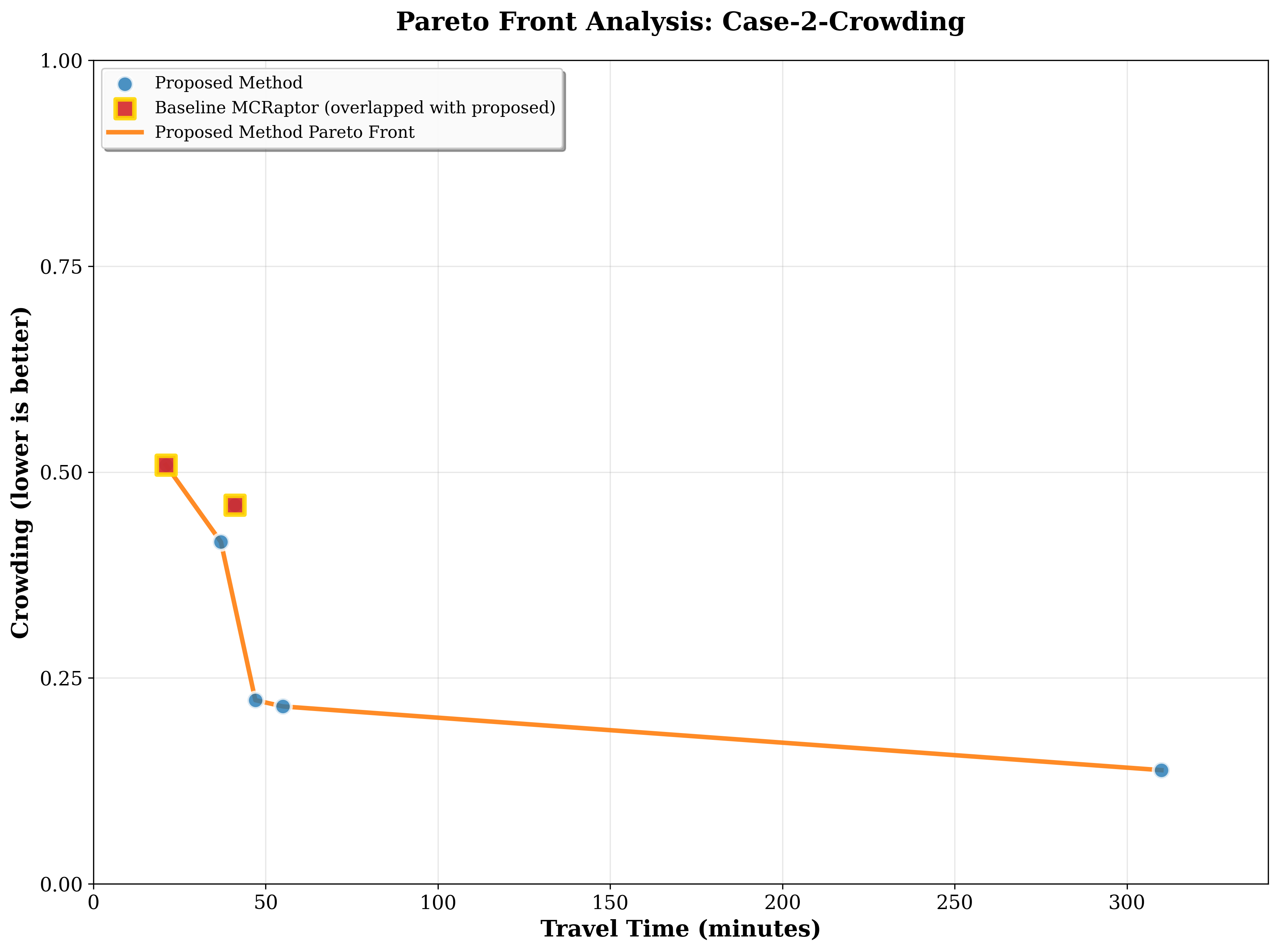}
        \caption{Case B: Pareto front in travel time vs crowding space, showing how ChatPlanner enlarges the solution space along the crowding dimension.}
        \label{fig:case2_pareto_crowding}
    \end{subfigure}
    
    \caption{Case B: Pareto front comparisons for ChatPlanner.}
    \label{fig:case2_pareto_combined} 
\end{figure}

\begin{figure}[htpb]
    \centering 
    
    \begin{subfigure}{0.495\textwidth}
        \centering
        \includegraphics[width=\textwidth]{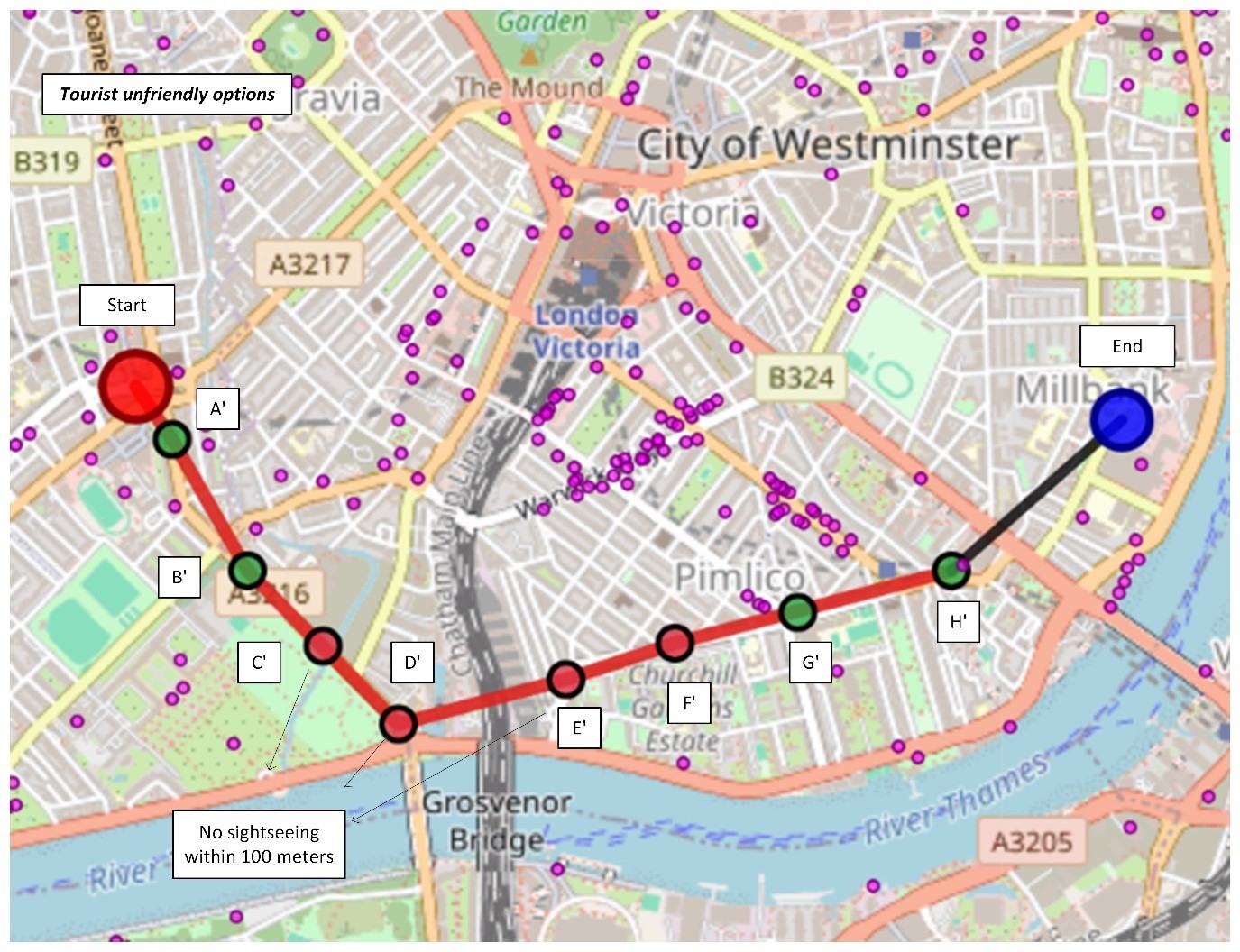}
        \caption{Case C: Route Planning with Sightseeing Need}
        \label{fig:case3_sol1}
    \end{subfigure}
    \hfill 
    \begin{subfigure}{0.495\textwidth}
        \centering
        \includegraphics[width=\textwidth]{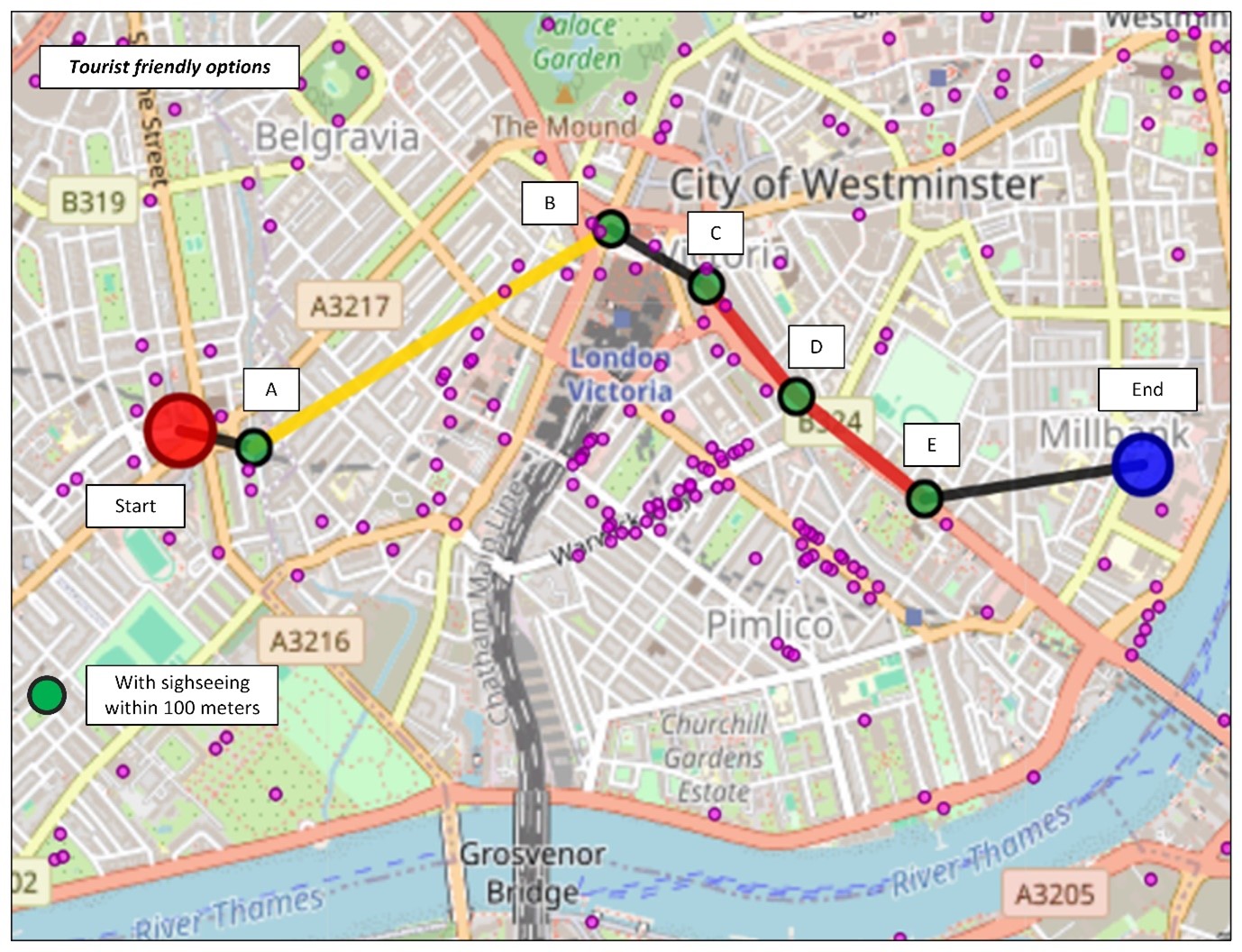}
        \caption{Case C: Route Planning with Sightseeing Need}
        \label{fig:case3_sol2}
    \end{subfigure}
    
    \caption{Case C: Route Planning with Sightseeing Need (two solutions).}
    \label{fig:case3_sols_combined} 
\end{figure}

\paragraph{Case C: Sightseeing Opportunities for Tourists}
In Case C, the journey goes from Sloane Square to Tate Britain by 10:00 for a tourist seeking interesting architecture. Figures~\ref{fig:case3_sol1} and \ref{fig:case3_sol2} compare different routing approaches. The first figure shows a direct route that passes through segments (A'–F') with few sightseeing opportunities within 100 meters, resulting in low sightseeing scores along the route. The second figure demonstrates ChatPlanner's tourist-friendly alternative that prioritizes points of interest and areas with high simulated sightseeing scores, even though it requires one transfer. This route achieves higher sightseeing scores along segments (A–E) while still efficiently connecting the same start and end locations.




\begin{figure}[htpb]
    \centering 
    
    \begin{subfigure}{0.495\textwidth}
        \centering
        \includegraphics[width=\textwidth]{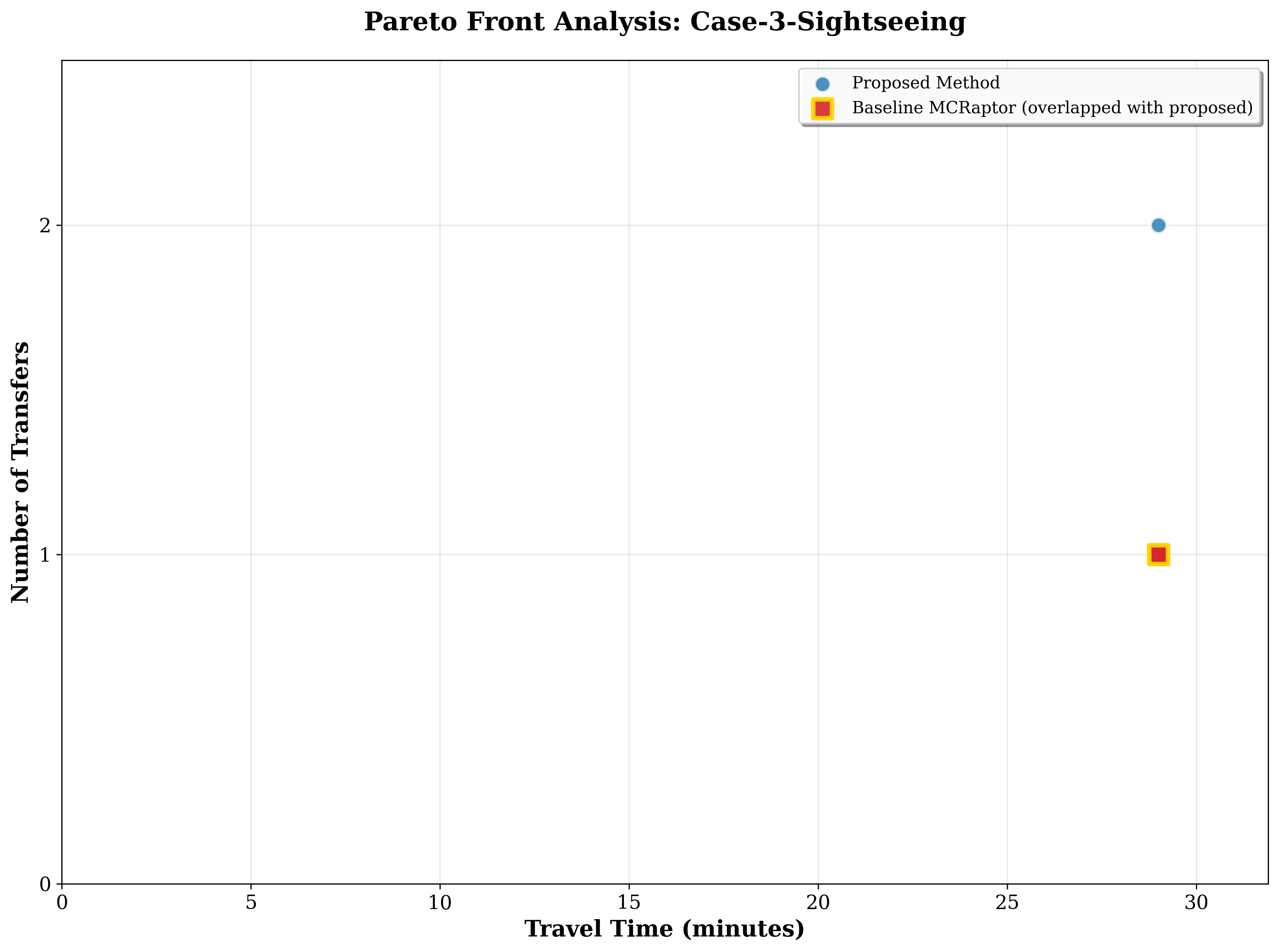}
        \caption{Case C: Pareto front comparison between ChatPlanner and MC-RAPTOR baseline in travel time vs number of transfers space.}
        \label{fig:case3_pareto_baseline}
    \end{subfigure}
    \hfill 
    \begin{subfigure}{0.495\textwidth}
        \centering
        \includegraphics[width=\textwidth]{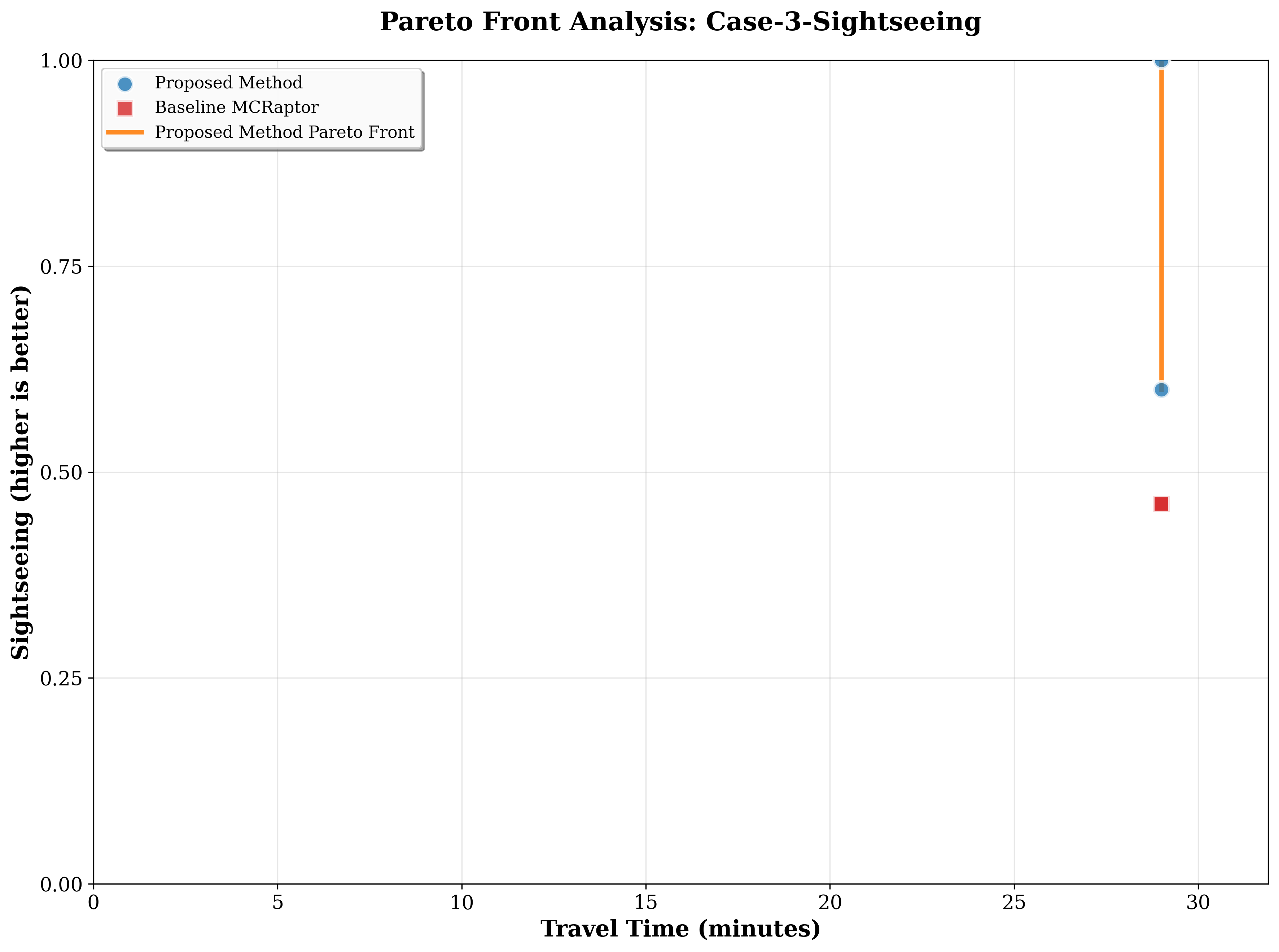}
        \caption{Case C: Pareto front in travel time vs sightseeing space, showing how ChatPlanner enlarges the solution space along the sightseeing dimension.}
        \label{fig:case3_pareto_sightseeing}
    \end{subfigure}
    
    \caption{Case C: Pareto front comparisons for ChatPlanner.}
    \label{fig:case3_pareto_combined} 
\end{figure}

Across all three cases, ChatPlanner successfully discovers and preserves personalized solutions that address specific user needs. Figures~\ref{fig:case1_pareto_combined}, \ref{fig:case2_pareto_combined}, and \ref{fig:case3_pareto_combined} present the Pareto front comparisons, showing that ChatPlanner expands the solution set along preference-specific dimensions (accessibility, crowding, sightseeing) compared to the baseline MC-RAPTOR algorithm. These results confirm that the framework effectively helps travelers customize their journeys.

\subsubsection{Quantitative Analysis of Multi-objective performance} \label{sec:quant_metrics_obj_space}
While visual inspections demonstrate ChatPlanner's ability to accommodate specific user preferences, this study provides a quantitative evaluation of the performance of the solutions to statistically demonstrate its advantage over the baseline. 

To evaluate across different travel scales, this study randomly sampled a subset of 30 journey requests to form our Routing Test Set. To reflect the spatial distribution of the transit network, these cases are divided into three distance categories based on the direct distance between the origin and destination, leading to 10 short journeys ($<5$ km), 10 medium journeys ($5$-$10$ km), and 10 long journeys ($>10$ km).

The baseline MC-RAPTOR algorithm optimizes only two objectives: travel time and the number of transfers. In contrast, ChatPlanner dynamically incorporates additional user-specific preferences as objectives. Post-hoc evaluation can be applied to MC-RAPTOR, allowing their solution sets to be analyzed and compared within a unified 6-dimensional objective space.

This study compares the Pareto fronts using three standard multi-objective performance indicators \citep{audet2021performance}: the Coverage metric (C-metric) \citep{zitzler1999multiobjective}, Inverted Generational Distance (IGD) \citep{zhang2007moea}, and Hypervolume (HV) \citep{zitzler1999multiobjective}. The results across 30 journey cases are summarized in Table~\ref{tab:pareto_comparison}.

\textbf{C-metric:} This metric measures the proportion of solutions in one set that are dominated by at least one solution in another set. 
$C(A,B)=1$ means $A$ dominates all solutions in $B$, whereas $C(A,B)=0$ means $A$ dominates none. As shown in Table~\ref{tab:pareto_comparison}, $C(\text{RAPTOR}, \text{ChatPlanner}) = 0.00$ across all 30 cases. This indicates that the baseline algorithm never dominates any solution generated by ChatPlanner in the 6D space. Conversely, $C(\text{ChatPlanner}, \text{RAPTOR})$ is greater than zero in several cases (mean = 0.25), demonstrating that ChatPlanner solutions frequently and strictly dominate the baseline solutions.


\textbf{IGD} measures the distance from a reference Pareto front to the algorithm's approximation set, evaluating both convergence and coverage. Since the true Pareto front is unknown, we construct the reference set as the union of non-dominated solutions from both MC-RAPTOR and ChatPlanner. ChatPlanner achieves a near-perfect mean IGD of 0.02, indicating that its solution set essentially constitutes the entire reference front. In contrast, the baseline algorithm yields a mean IGD of 0.86, remaining far from the true multi-objective front due to its exclusive focus on time and transfers.



\textbf{HV} measures the volume of the objective space dominated by the solution set; higher values indicate better diversity and optimality. In the full 6-dimensional objective space, ChatPlanner achieves higher hypervolume than the baseline in 28 of the 29 cases where HV is defined, with a mean of 0.40 compared to 0.08 for the baseline. In some cases, this improvement is considerable, with ChatPlanner achieving substantially higher hypervolume where the baseline performs poorly. This result directly reflects ChatPlanner’s ability to explore multiple user-specific criteria that MC-RAPTOR does not, producing solutions that collectively dominate a much larger region of the objective space.

\begin{table}[htpb]
\centering
\caption{Per-case comparison of MC-RAPTOR (R) and ChatPlanner (CP). HV: per-case min-max normalized objectives, reference point 1.1. ``--'': single shared route, HV undefined.}
\label{tab:pareto_comparison}
\small
\setlength{\tabcolsep}{4pt}
\renewcommand{\arraystretch}{1.025}

\begin{tabular}{@{} l r r r r r r r r @{}}
\toprule
& \multicolumn{2}{c}{Pareto size $|\mathcal{P}|$}
& \multicolumn{2}{c}{C-metric}
& \multicolumn{2}{c}{IGD $\downarrow$}
& \multicolumn{2}{c}{HV $\uparrow$} \\
\cmidrule(lr){2-3}
\cmidrule(lr){4-5}
\cmidrule(lr){6-7}
\cmidrule(lr){8-9}
Case
& R & CP
& $C(\text{CP,R}){\uparrow}$ & $C(\text{R,CP}){\downarrow}$
& R & CP & R & CP \\
\midrule
\multicolumn{9}{@{}l}{\textit{Long journeys}} \\[1pt]
long\_1   & 1 &  22 & 0.00 & 0.00 & 1.49 & 0.03 & 0.01 & 0.69 \\
long\_2   & 2 &  16 & 0.00 & 0.00 & 0.71 & 0.00 & 0.04 & 0.13 \\
long\_3   & 1 &   3 & 0.00 & 0.00 & 1.00 & 0.00 & 0.01 & 0.02 \\
long\_4   & 2 &  23 & 1.00 & 0.00 & 0.87 & 0.00 & 0.05 & 0.23 \\
long\_5   & 1 &   5 & 0.00 & 0.00 & 1.05 & 0.00 & 0.01 & 0.05 \\
long\_6   & 1 &  17 & 0.00 & 0.00 & 1.17 & 0.00 & 0.04 & 0.44 \\
long\_7   & 2 &  68 & 1.00 & 0.00 & 0.73 & 0.00 & 0.07 & 0.74 \\
long\_8   & 1 &   8 & 1.00 & 0.00 & 1.34 & 0.00 & 0.01 & 0.85 \\
long\_9   & 2 &  13 & 0.00 & 0.00 & 0.58 & 0.00 & 0.11 & 0.42 \\
long\_10  & 2 &  47 & 1.00 & 0.00 & 0.89 & 0.00 & 0.06 & 0.42 \\
\midrule
\multicolumn{9}{@{}l}{\textit{Medium journeys}} \\[1pt]
medium\_1  & 2 &  3 & 0.50 & 0.00 & 0.53 & 0.00 & 0.01 & 0.14 \\
medium\_2  & 2 & 11 & 1.00 & 0.00 & 0.88 & 0.00 & 0.16 & 0.29 \\
medium\_3  & 2 & 19 & 0.00 & 0.00 & 0.82 & 0.00 & 0.05 & 0.72 \\
medium\_4  & 1 & 29 & 0.00 & 0.00 & 1.16 & 0.00 & 0.04 & 0.50 \\
medium\_5  & 2 &  6 & 0.00 & 0.00 & 0.65 & 0.10 & 0.02 & 0.26 \\
medium\_6  & 1 & 28 & 0.00 & 0.00 & 1.22 & 0.00 & 0.01 & 0.57 \\
medium\_7  & 1 &  3 & 0.00 & 0.00 & 0.98 & 0.00 & 0.00 & 0.02 \\
medium\_8  & 2 & 23 & 0.00 & 0.00 & 0.79 & 0.02 & 0.08 & 0.60 \\
medium\_9  & 2 &  5 & 0.00 & 0.00 & 0.46 & 0.00 & 0.02 & 0.03 \\
medium\_10 & 2 &  6 & 0.00 & 0.00 & 0.70 & 0.08 & 0.05 & 0.05 \\
\midrule
\multicolumn{9}{@{}l}{\textit{Short journeys}} \\[1pt]
short\_1  & 2 & 33 & 0.00 & 0.00 & 0.98 & 0.02 & 0.26 & 0.56 \\
short\_2  & 1 & 53 & 0.00 & 0.00 & 1.08 & 0.00 & 0.72 & 1.33 \\
short\_3  & 1 &  8 & 0.00 & 0.00 & 0.96 & 0.00 & 0.01 & 0.46 \\
short\_4  & 1 &  6 & 1.00 & 0.00 & 1.30 & 0.00 & 0.01 & 0.16 \\
short\_5  & 2 &  4 & 0.00 & 0.00 & 0.64 & 0.18 & 0.10 & 0.14 \\
short\_6  & 2 &  5 & 0.00 & 0.00 & 0.74 & 0.13 & 0.12 & 0.13 \\
short\_7  & 2 & 12 & 0.00 & 0.00 & 0.78 & 0.09 & 0.04 & 0.68 \\
short\_8  & 2 &  5 & 0.50 & 0.00 & 0.54 & 0.00 & 0.17 & 0.54 \\
short\_9  & 2 & 18 & 0.50 & 0.00 & 0.65 & 0.00 & 0.01 & 0.36 \\
short\_10 & 1 &  1 & 0.00 & 0.00 & 0.00 & 0.00 & - & - \\
\midrule
\multicolumn{9}{@{}l}{\textit{Summary (30 cases)}} \\[1pt]
Mean   & 1.60 & 16.67 & 0.25 & 0.00 & 0.86 & 0.02 & 0.08 & 0.40 \\
Std    & 0.50 & 16.20 & 0.41 & 0.00 & 0.31 & 0.05 & 0.14 & 0.31 \\
Median & 2.00 & 11.50 & 0.00 & 0.00 & 0.84 & 0.00 & 0.04 & 0.42 \\
\bottomrule
\end{tabular}

\end{table}

\begin{table}[htpb]
\small
\centering
\caption{Average language processing time per user request across 30 test cases.}
\label{tab:llm_latency}
\setlength{\tabcolsep}{6pt} 
\renewcommand{\arraystretch}{1.15}

\begin{tabular}{l c c c c}
\toprule
\textbf{Model Configuration} & \textbf{Mean (s)} & \textbf{Std Dev} & \textbf{Median} & \textbf{90th Percentile} \\
\midrule
Qwen Base & 4.84 & 0.12 & 4.82 & 4.98 \\
Qwen Base + RAG & 5.14 & 0.20 & 5.14 & 5.31 \\
Llama Base & 5.30 & 0.31 & 5.22 & 5.44 \\
Llama Base + RAG & 5.80 & 0.16 & 5.80 & 6.01 \\
Qwen Fine-tuned & 4.77 & 0.24 & 4.76 & 4.89 \\
Llama Fine-tuned & 5.28 & 0.26 & 5.29 & 5.45 \\
Qwen Fine-tuned + RAG & 5.32 & 0.22 & 5.34 & 5.46 \\
Llama Fine-tuned + RAG & \textbf{5.79} & \textbf{0.23} & \textbf{5.80} & \textbf{5.91} \\
\bottomrule
\end{tabular}

\end{table}

\subsection{Experiment 4: Latency and Computational Tractability}\label{sec:exp4_tractability}

To assess the viability of ChatPlanner for real-world deployment, the computation time of both blocks was evaluated for latency and computational tractability. The ChatPlanner pipeline consists of two main computational blocks in the initial planning phase: the natural language processing block, where the LLM interprets the user's request, and the public transit routing block, where the MC-RAPTOR algorithm calculates paths.

\subsubsection{Language Processing Latency}

First, this experiment measured the time required for the LLM to process a user's natural language travel request. This includes retrieving similar examples from the database (RAG), processing the text, and generating the structured output with the interpreted preference scores. 30 user requests were randomly sampled from the User Study Test Set (described in Section~\ref{sec:exp2_setup}) and measured the processing time on our server with an NVIDIA A40 GPU for different model configurations.

As shown in Table~\ref{tab:llm_latency}, our best-performing configuration from Experiment 2 (Llama Fine-tuned + RAG) takes an average of 5.79 seconds to process requests. This processing time is considered reasonable and acceptable, as it falls within the established 10-second usability threshold for keeping a user's attention focused on a continuous dialogue task \citep{nielsen1994usability}. Furthermore, this running time can still be optimized significantly in real-world product deployment settings through hardware scaling and server-side memory management optimizations, such as continuous batching and PagedAttention \citep{kwon2023efficient}.

\subsubsection{Route Search Latency and Computational Tractability}
While the language processing takes about 6 seconds, the majority of the system's total response time is determined by the public transit routing algorithm. MC-RAPTOR is highly sensitive to the number of objectives considered. The algorithm tracks and compares non-dominated solutions at every transit stop. Therefore, increasing the number of objectives causes an exponential increase in calculation time.

To manage this computational burden, a core design choice in the ChatPlanner framework is the preference selection mechanism in Node 3, which restricts the number of active user-defined criteria to $K=2$. To provide a principled computational justification for this choice, the sensitivity of the MC-RAPTOR algorithm to the number of simultaneous objectives is analyzed. Specifically, experiments are conducted by incrementally adding user preferences (crowdedness, accessibility, safety, and sightseeing) to the base routing objectives (travel time and number of transfers). The average execution time is measured across the same 30 journey cases evaluated in Section~\ref{sec:quant_metrics_obj_space} (10 short, 10 medium, and 10 long journeys). As shown in Table~\ref{tab:computation_time_k}, a standard routing search (2 objectives) requires an average of 12.6 seconds. Adding two additional user preferences increases the average search time to around 50 seconds. However, expanding the number of objectives to 5 causes an immediate combinatorial explosion, pushing average computation times past 8 minutes. Attempting to optimize all 6 criteria simultaneously results in complete algorithm failure, exhausting the server's 86\,GB of system memory on long and medium journeys.


\subsubsection{Overall Deployment Feasibility}
When combining the LLM processing time and the MC-RAPTOR search time for a $K=2$ configuration, ChatPlanner requires an average end-to-end processing time of approximately 1 minute to generate its initial response.

Compared to standard routing engines such as Google Maps, which primarily optimize only for travel time and compute paths in a few seconds, ChatPlanner's latency is notably higher. However, ChatPlanner acts as a personalized travel assistant, solving a highly complex four-objective optimization problem tailored specifically to users' preferences. Research on human-chatbot interaction suggests that users tolerate higher response times and even report increased trust when the system is perceived to be performing complex, personalized computation \citep{gnewuch2018faster, buell2011labor}. Also, the current implementation is not optimized for running speed, and further reductions are achievable through code optimization, hardware scaling, and parallelizing independent route traversals across multiple cores \citep{delling2015round}. Importantly, the modular design of the ChatPlanner framework allows the routing solver to be replaced with any faster multi-objective algorithm without modifying the rest of the pipeline.

\begin{table}[htpb]
\small
\centering
\caption{Sensitivity analysis of MC-RAPTOR computation time relative to the number of simultaneous optimization objectives.}
\label{tab:computation_time_k}
\setlength{\tabcolsep}{4pt}
\renewcommand{\arraystretch}{1.15}
\begin{tabularx}{\textwidth}{L c c c c c}
\toprule
\makecell{\scriptsize\textbf{Algorithm Configuration}} &
\makecell{\scriptsize\textbf{Total}\\\scriptsize\textbf{Objectives}} &
\makecell{\scriptsize\textbf{Long Journeys}\\\scriptsize\textbf{($>10$km) [s]}} &
\makecell{\scriptsize\textbf{Medium Journeys}\\\scriptsize\textbf{(5--10km) [s]}} &
\makecell{\scriptsize\textbf{Short Journeys}\\\scriptsize\textbf{($<5$km) [s]}} &
\makecell{\scriptsize\textbf{Average}\\\scriptsize\textbf{Time [s]}} \\
\midrule
Base (Time, Transfers) ($K=0$) & 2 & 17.32 & 16.26 & 4.12 & 12.57 \\
Base + 1 Preference ($K=1$) & 3 & 43.69 & 42.11 & 41.89 & 42.56 \\
ChatPlanner Initial Max ($K=2$) & 4 & 64.83 & 43.25 & 43.16 & 50.41 \\
Base + 3 Preferences ($K=3$) & 5 & 907.27 & 457.91 & 197.73 & 520.97 \\
Base + 4 Preferences ($K=4$) & 6 & \textit{$-$} & \textit{$-$} & 857.79 & \textit{$-$} \\
\bottomrule
\end{tabularx}
\end{table}



\newpage

\section{Conclusions}
\label{sec:conclusions}


This research presents ChatPlanner, a novel framework for applying LLMs to public transit routing, addressing the critical gap in personalized route planning. This approach integrates LLM capabilities into the public transit routing algorithm to capture users' conversationally expressed preferences as preference scores alongside traditional routing inputs extracted from natural language queries, such as origin, destination, and requested arrival time. These extracted preferences guide the selection of criteria, which are then incorporated into the objective function of the public transit routing algorithm, enabling the generation of personalized, preference-aware solution sets.


Four experiments are conducted to validate the framework: solutions' feasibility compared with a tool-augmented LLM, extraction of routing information and preferences, solution set quality and completeness, and system latency and computational tractability.



Comparative analysis reveals that even tool-augmented LLMs, such as GPT-5.0 equipped with Google Maps API, Moovit, and Rome2Rio, cannot stably perform end-to-end public transit routing, achieving only one-third feasibility rate, while ChatPlanner generates feasible solutions reliably. These limitations of tool-augmented LLMs stem from the inherent instability, as the model cannot always guarantee successful API calls. The model cannot reliably resolve ambiguous location names across different external databases and could misinterpret transit information when parsing unstructured web layouts. The model can also misreport travel durations and walking segments when summarizing route details from external sources. ChatPlanner addresses these constraints by strictly limiting the LLM to natural language parsing and delegating the path planning to a routing algorithm, with both components sharing the same database of stop names and transit network data.


Testing on an out-of-distribution, human-written dataset, our evaluation across three user query extraction tasks validates the necessity of combining fine-tuning with RAG. Fine-tuning with our designed input-output pairs is essential for extracting origin, destination, and requested arrival time from natural language queries, while RAG is needed to provide query-specific context that resolves imprecise, conversational human expressions for locations and times. For preference interpretation, datasets incorporating eight personas and five contexts establish scoring standards for both fine-tuning and RAG components. The combination achieves better rubric-consistent preference interpretation compared to either component alone. Fine-tuning learns preference patterns, and RAG provides query-specific examples that guide the model toward consistent and accurate scoring aligned with the designed standards.


Our case studies demonstrate that by scoring users' expressed preferences, ChatPlanner identifies preference-relevant solutions across different criteria dimensions that existing route planners overlook. From a user-centric perspective, ChatPlanner provides a richer set of route alternatives. From an algorithmic standpoint, this demonstrates how distinct objective functions operating in different search regions of the solution space provide differentiated Pareto frontiers, thereby expanding the solution set for personalized transit routing.

The latency evaluation shows that ChatPlanner generates its initial response in about one minute end to end, and that limiting the number of active preferences in Node 3 is necessary to keep the route search tractable. While this latency is higher than that of standard routing engines, the implementation is not yet optimized for speed, and the modular design allows a faster routing solver to be substituted.


\section{Future Directions}
\label{sec:future_directions}

Looking ahead, the integration of LLMs in public transit systems warrants further investigation in several key areas. First, enhanced solution algorithms could provide users with a broader range of preference-relevant routing alternatives. If alternative route planners employing different search mechanisms, such as evolutionary algorithms, can generate more diverse solution sets, the current ChatPlanner framework could similarly benefit from incorporating these approaches to expand its solution space. Second, replacing the simulated crowding and safety levels and the distance-based sightseeing rule with empirical datasets or integrating with corresponding models would enable validated, real-world personalization across diverse traveler scenarios, from rush-hour commuters to night-time travelers and tourists. Third, improving preference interpretation accuracy remains a critical research direction. LLMs trained on general text may associate words in ways that do not match their meaning in a transportation context. Developing techniques to strengthen domain-specific understanding of transit terminology represents a promising avenue for more reliable LLM-based transit planning systems.

\section*{Acknowledgments}
The authors would like to thank [Chinese Scholarship Council] for their support and contributions to this research.

\newpage
\bibliographystyle{unsrtnat}
\bibliography{references}

\newpage
\appendix

\section{Empirical Justification of the Activation Threshold}
\label{apx:threshold_justification}

To determine the appropriate threshold for converting continuous LLM preference predictions into active routing criteria (Node 3), we analyzed the default output behavior of the models. We evaluated 30 semantically neutral journey requests. These requests contained only basic travel information (origin, destination, and arrival time) without any explicit or implicit preference keywords. An example of such a request is: \textit{``I need to get from Temple Fortune Lane to Suffolk Road by 10:54.''}

Table~\ref{tab:neutral_sentence_scores} presents the average preference scores predicted by different model configurations across these 30 neutral sentences. The results show that LLMs do not output scores of zero for neutral inputs. Base models often generate inaccurately high scores, such as Llama Base predicting an average safety score of 0.961. 

While the combination of fine-tuning and RAG successfully reduces these extreme values, the average predicted scores stabilize around 0.5 rather than 0.0. Because the default output for a neutral request naturally centers around 0.5, setting an activation threshold is a necessary structural requirement to separate the LLM's default output values from actively expressed travel preferences. This threshold is set to 0.6 to ensure the route planning algorithm only optimizes for criteria the user has actively expressed.

\begin{table}[htpb]
\small
\centering
\caption{Average LLM preference score predictions across 30 semantically neutral test sentences.}
\label{tab:neutral_sentence_scores}
\begin{tabular}{l c c c c}
\toprule
\textbf{Model Setup} & \textbf{Accessibility} & \textbf{Crowdedness} & \textbf{Safety} & \textbf{Sightseeing} \\
\midrule
Llama Base & 0.835 & 0.427 & 0.961 & 0.615 \\
Qwen Base & 0.900 & 0.676 & 0.948 & 0.250 \\
Llama Base + RAG & 0.676 & 0.416 & 0.510 & 0.046 \\
Qwen Base + RAG & 0.610 & 0.623 & 0.630 & 0.020 \\
\midrule
Llama FT & 0.526 & 0.559 & 0.578 & 0.477 \\
Qwen FT & 0.544 & 0.544 & 0.576 & 0.441 \\
Llama FT + RAG & \textbf{0.546} & \textbf{0.533} & \textbf{0.486} & \textbf{0.516} \\
Qwen FT + RAG & \textbf{0.544} & \textbf{0.526} & \textbf{0.483} & \textbf{0.510} \\
\bottomrule
\end{tabular}
\end{table}

\section{Sensitivity of the Node-3 Gate: Downstream Effect on the Routing Outputs}
\label{apx:sensitivity_txk}

To measure how the gate settings in Node~3 affect the actual routing outputs, we run the full $K\times T$ grid ($K\in\{0,1,2,3\}$, $T\in\{0.4,\dots,0.8\}$; 16 settings as $K=0$ is $T$-independent; deployed default $D$: $K{=}2$, $T{=}0.6$) on 74 of the 78 cases of the User Route Choice Set (Section~\ref{sec:route-pref-study}); the remaining four are excluded because their $K{=}3$ solve did not finish within 10 hours. The Node-1 parses are computed once per case and frozen across all settings, so the gate is the only variable. The effect of the gate depends on how many of a case's predicted scores are high enough to activate: if no score is even above the lowest threshold, all 16 settings behave identically. Therefore, the cases are grouped by the number of preferences active at the most permissive setting ($T{=}0.4$): 3-active (21 cases), 2-active (31), 1-active (17), and 0-active (5). For example, a case with predicted scores $0.72$, $0.65$, $0.45$, $0.30$ is 3-active.

For each case and setting, the route planner optimizes travel time, number of transfers, and the active preferences as objectives. For each case, Tables~\ref{tab:kt-grid-3active}--\ref{tab:kt-grid-0active} report two quantities.

\begin{itemize}
    \item The Pareto-front size $|P|$: the number of non-dominated routes a setting returns, shown as a $4\times5$ mini-grid per case (rows $K{=}0$-$3$, columns $T{=}0.8$ down to $0.4$; the default's cell in bold italics).
    \item The default purity $\pi_D$, one value per case: we pool the routes of all 16 settings, extract the non-dominated set of this union (the reference front), and report the fraction of the default's routes that lie on it. Post-hoc evaluation is applied to every route in different settings, allowing the solutions to be compared within the same objective space.
\end{itemize}

The results show that the gate settings do change the routing outputs, with the effect growing with the number of active preferences: nothing changes in the 0-active group, only the threshold matters in the 1-active group, and in the 2- and 3-active groups tightening the gate (higher $T$, lower $K$) shrinks the Pareto front while relaxing the gate enlarges the Pareto front, and in the most extreme case from 21 routes at the default to 162 at $K{=}3$ (G3-C18). The purity values quantify how much of the default's output survives the wider search, where a value below $1.00$ means that some of the default's routes are dominated by routes found under other settings. The default keeps full purity ($\pi_D{=}1.00$) in 59 of 74 cases. The remaining 15 cases stay moderate to high, with a median of $0.79$ and a minimum of $0.67$ (G3-C12, G3-C16, G3-C20); 13 of them are 3-active cases where a third preference is capped. This sensitivity analysis can serve as a reference for configuring the threshold $T$ and the top-$K$ cap.


\providecommand{\ktgrid}[4]{{\scriptsize\setlength{\tabcolsep}{1.5pt}\renewcommand{\arraystretch}{0.85}\begin{tabular}{@{}c|ccccc@{}}$K\backslash T$ & $0.8$ & $0.7$ & $0.6$ & $0.5$ & $0.4$ \\ \cmidrule(lr){1-6}$0$ & \multicolumn{5}{c}{#1} \\ $1$ & #2 \\ $2$ & #3 \\ $3$ & #4 \\ \end{tabular}}}

{\footnotesize\setlength{\tabcolsep}{3pt}
\begin{longtable}{@{}l l c c @{\hspace{4pt}}|@{\hspace{4pt}} l l c c@{}}
\caption{Per-case $K\times T$ Pareto grid, \textbf{3-active} group (21 cases, two per row). The $|P|$ column is the $4$-$K\times5$-$T$ mini-grid (row labels $K$, column header $T$), showing the Pareto-front size at every setting; the default ($K{=}2$, $T{=}0.6$) value is in \textbf{\textit{bold italics}}. \emph{Active prefs} lists the case's preferences active at $T{=}0.4$.}\label{tab:kt-grid-3active}\\
\toprule
Case & Active prefs & $|P|$ & $\pi_D$ & Case & Active prefs & $|P|$ & $\pi_D$ \\
\midrule \endfirsthead
\multicolumn{8}{@{}l@{}}{\scriptsize\emph{Table~\ref{tab:kt-grid-3active} (continued).} Each $|P|$ cell is a $4$-$K\times5$-$T$ grid.}\\[2pt]
\toprule
Case & Active prefs & $|P|$ & $\pi_D$ & Case & Active prefs & $|P|$ & $\pi_D$ \\
\midrule \endhead
\midrule \multicolumn{8}{r@{}}{\footnotesize continued on next page}\\ \endfoot
\bottomrule \endlastfoot
G3-C1 & cro, acc, saf & \ktgrid{2}{3&3&3&3&3}{4&4&\textbf{\textit{4}}&4&4}{4&24&24&24&24} & 1.00 & G3-C2 & cro, acc, saf & \ktgrid{1}{10&10&10&10&10}{10&19&\textbf{\textit{19}}&19&19}{10&19&19&67&67} & 0.84 \\
\addlinespace[2pt]\cmidrule(l{1pt}r{1pt}){1-4}\cmidrule(l{1pt}r{1pt}){5-8}
G3-C3 & acc, cro, saf & \ktgrid{2}{2&2&2&5&5}{2&2&\textbf{\textit{2}}&21&21}{2&2&2&79&79} & 1.00 & G3-C4 & acc, cro, saf & \ktgrid{1}{1&1&1&1&1}{6&6&\textbf{\textit{6}}&6&6}{6&6&6&35&35} & 0.83 \\
\addlinespace[2pt]\cmidrule(l{1pt}r{1pt}){1-4}\cmidrule(l{1pt}r{1pt}){5-8}
G3-C5 & acc, cro, saf & \ktgrid{1}{1&1&1&1&1}{1&6&\textbf{\textit{6}}&6&6}{1&6&6&17&17} & 0.83 & G3-C6 & cro, saf, acc & \ktgrid{1}{2&2&2&2&2}{2&55&\textbf{\textit{55}}&55&55}{2&55&55&55&83} & 1.00 \\
\addlinespace[2pt]\cmidrule(l{1pt}r{1pt}){1-4}\cmidrule(l{1pt}r{1pt}){5-8}
G3-C7 & cro, saf, acc & \ktgrid{1}{2&2&2&2&2}{2&55&\textbf{\textit{55}}&55&55}{2&55&83&83&83} & 1.00 & G3-C8 & saf, cro, acc & \ktgrid{2}{6&6&6&6&6}{6&25&\textbf{\textit{25}}&25&25}{6&25&32&32&32} & 1.00 \\
\addlinespace[2pt]\cmidrule(l{1pt}r{1pt}){1-4}\cmidrule(l{1pt}r{1pt}){5-8}
G3-C9 & cro, saf, sig & \ktgrid{1}{12&12&12&12&12}{12&38&\textbf{\textit{38}}&38&38}{12&38&38&75&75} & 0.79 & G3-C10 & acc, saf, sig & \ktgrid{2}{2&2&2&2&2}{14&14&\textbf{\textit{14}}&14&14}{26&26&26&26&26} & 1.00 \\
\addlinespace[2pt]\cmidrule(l{1pt}r{1pt}){1-4}\cmidrule(l{1pt}r{1pt}){5-8}
G3-C11 & saf, sig, acc & \ktgrid{2}{5&5&5&5&5}{6&6&\textbf{\textit{6}}&6&6}{6&6&6&9&9} & 1.00 & G3-C12 & saf, acc, cro & \ktgrid{2}{3&3&3&3&3}{3&3&\textbf{\textit{3}}&3&3}{3&3&3&13&13} & 0.67 \\
\addlinespace[2pt]\cmidrule(l{1pt}r{1pt}){1-4}\cmidrule(l{1pt}r{1pt}){5-8}
G3-C13 & cro, saf, acc & \ktgrid{1}{8&8&8&8&8}{66&66&\textbf{\textit{66}}&66&66}{66&66&66&107&107} & 0.97 & G3-C14 & cro, sig, saf & \ktgrid{1}{7&7&7&7&7}{16&16&\textbf{\textit{16}}&16&16}{16&16&16&46&46} & 0.75 \\
\addlinespace[2pt]\cmidrule(l{1pt}r{1pt}){1-4}\cmidrule(l{1pt}r{1pt}){5-8}
G3-C15 & sig, cro, saf & \ktgrid{2}{2&2&2&2&2}{2&18&\textbf{\textit{18}}&18&18}{2&18&88&88&88} & 0.78 & G3-C16 & saf, acc, cro & \ktgrid{1}{3&3&3&3&3}{3&3&\textbf{\textit{3}}&3&3}{3&3&3&5&5} & 0.67 \\
\addlinespace[2pt]\cmidrule(l{1pt}r{1pt}){1-4}\cmidrule(l{1pt}r{1pt}){5-8}
G3-C17 & acc, saf, cro & \ktgrid{2}{5&5&5&5&5}{13&13&\textbf{\textit{13}}&13&13}{13&13&33&33&33} & 0.85 & G3-C18 & sig, cro, saf & \ktgrid{1}{4&4&4&4&4}{4&21&\textbf{\textit{21}}&21&21}{4&162&162&162&162} & 0.86 \\
\addlinespace[2pt]\cmidrule(l{1pt}r{1pt}){1-4}\cmidrule(l{1pt}r{1pt}){5-8}
G3-C19 & acc, cro, saf & \ktgrid{2}{3&3&3&3&3}{3&3&\textbf{\textit{3}}&4&4}{3&3&3&11&11} & 1.00 & G3-C20 & acc, saf, cro & \ktgrid{2}{4&4&4&4&4}{9&9&\textbf{\textit{9}}&9&9}{9&9&9&27&27} & 0.67 \\
\addlinespace[2pt]\cmidrule(l{1pt}r{1pt}){1-4}\cmidrule(l{1pt}r{1pt}){5-8}
G3-C21 & saf, acc, cro & \ktgrid{2}{4&4&4&4&4}{10&10&\textbf{\textit{10}}&10&10}{10&10&57&57&57} & 0.70 &  & & &  \\
\addlinespace[2pt]
\end{longtable}
}

\providecommand{\ktgrid}[4]{{\scriptsize\setlength{\tabcolsep}{1.5pt}\renewcommand{\arraystretch}{0.85}\begin{tabular}{@{}c|ccccc@{}}$K\backslash T$ & $0.8$ & $0.7$ & $0.6$ & $0.5$ & $0.4$ \\ \cmidrule(lr){1-6}$0$ & \multicolumn{5}{c}{#1} \\ $1$ & #2 \\ $2$ & #3 \\ $3$ & #4 \\ \end{tabular}}}

{\footnotesize\setlength{\tabcolsep}{3pt}
\begin{longtable}{@{}l l c c @{\hspace{4pt}}|@{\hspace{4pt}} l l c c@{}}
\caption{Per-case $K\times T$ Pareto grid, \textbf{2-active} group (31 cases). Layout, metrics and abbreviations as in Table~\ref{tab:kt-grid-3active}; because at most two preferences are active, $K{=}3$ reproduces $K{=}2$.}\label{tab:kt-grid-2active}\\
\toprule
Case & Active prefs & $|P|$ & $\pi_D$ & Case & Active prefs & $|P|$ & $\pi_D$ \\
\midrule \endfirsthead
\multicolumn{8}{@{}l@{}}{\scriptsize\emph{Table~\ref{tab:kt-grid-2active} (continued).} Each $|P|$ cell is a $4$-$K\times5$-$T$ grid.}\\[2pt]
\toprule
Case & Active prefs & $|P|$ & $\pi_D$ & Case & Active prefs & $|P|$ & $\pi_D$ \\
\midrule \endhead
\midrule \multicolumn{8}{r@{}}{\footnotesize continued on next page}\\ \endfoot
\bottomrule \endlastfoot
G2-C1 & saf, cro & \ktgrid{1}{1&3&3&3&3}{1&3&\textbf{\textit{17}}&17&17}{1&3&17&17&17} & 1.00 & G2-C2 & saf, cro & \ktgrid{1}{1&4&4&4&4}{1&4&\textbf{\textit{7}}&7&7}{1&4&7&7&7} & 1.00 \\
\addlinespace[2pt]\cmidrule(l{1pt}r{1pt}){1-4}\cmidrule(l{1pt}r{1pt}){5-8}
G2-C3 & cro, acc & \ktgrid{1}{8&8&8&8&8}{16&16&\textbf{\textit{16}}&16&16}{16&16&16&16&16} & 1.00 & G2-C4 & saf, sig & \ktgrid{1}{3&3&3&3&3}{5&5&\textbf{\textit{5}}&5&5}{5&5&5&5&5} & 1.00 \\
\addlinespace[2pt]\cmidrule(l{1pt}r{1pt}){1-4}\cmidrule(l{1pt}r{1pt}){5-8}
G2-C5 & saf, sig & \ktgrid{1}{9&9&9&9&9}{22&22&\textbf{\textit{22}}&22&22}{22&22&22&22&22} & 1.00 & G2-C6 & saf, cro & \ktgrid{2}{4&4&4&4&4}{4&4&\textbf{\textit{4}}&23&23}{4&4&4&23&23} & 1.00 \\
\addlinespace[2pt]\cmidrule(l{1pt}r{1pt}){1-4}\cmidrule(l{1pt}r{1pt}){5-8}
G2-C7 & saf, cro & \ktgrid{2}{8&8&8&8&8}{8&8&\textbf{\textit{8}}&84&84}{8&8&8&84&84} & 0.75 & G2-C8 & acc, cro & \ktgrid{2}{2&2&2&3&3}{2&2&\textbf{\textit{2}}&4&4}{2&2&2&4&4} & 1.00 \\
\addlinespace[2pt]\cmidrule(l{1pt}r{1pt}){1-4}\cmidrule(l{1pt}r{1pt}){5-8}
G2-C9 & cro, saf & \ktgrid{2}{7&7&7&7&7}{55&55&\textbf{\textit{55}}&55&55}{55&55&55&55&55} & 1.00 & G2-C10 & saf, cro & \ktgrid{2}{2&5&5&5&5}{2&5&\textbf{\textit{5}}&14&14}{2&5&5&14&14} & 0.80 \\
\addlinespace[2pt]\cmidrule(l{1pt}r{1pt}){1-4}\cmidrule(l{1pt}r{1pt}){5-8}
G2-C11 & cro, saf & \ktgrid{1}{1&5&5&5&5}{1&44&\textbf{\textit{44}}&44&44}{1&44&44&44&44} & 1.00 & G2-C12 & acc, cro & \ktgrid{2}{3&3&3&3&3}{16&16&\textbf{\textit{16}}&16&16}{16&16&16&16&16} & 1.00 \\
\addlinespace[2pt]\cmidrule(l{1pt}r{1pt}){1-4}\cmidrule(l{1pt}r{1pt}){5-8}
G2-C13 & acc, cro & \ktgrid{2}{2&2&2&2&2}{3&3&\textbf{\textit{3}}&3&3}{3&3&3&3&3} & 1.00 & G2-C14 & cro, saf & \ktgrid{2}{11&11&11&11&11}{85&85&\textbf{\textit{85}}&85&85}{85&85&85&85&85} & 1.00 \\
\addlinespace[2pt]\cmidrule(l{1pt}r{1pt}){1-4}\cmidrule(l{1pt}r{1pt}){5-8}
G2-C15 & acc, saf & \ktgrid{2}{2&2&2&2&2}{8&8&\textbf{\textit{8}}&8&8}{8&8&8&8&8} & 1.00 & G2-C16 & acc, saf & \ktgrid{1}{4&4&4&4&4}{14&14&\textbf{\textit{14}}&14&14}{14&14&14&14&14} & 1.00 \\
\addlinespace[2pt]\cmidrule(l{1pt}r{1pt}){1-4}\cmidrule(l{1pt}r{1pt}){5-8}
G2-C17 & sig, cro & \ktgrid{1}{6&6&6&6&6}{6&27&\textbf{\textit{27}}&27&27}{6&27&27&27&27} & 1.00 & G2-C18 & cro, acc & \ktgrid{1}{1&3&3&3&3}{1&3&\textbf{\textit{3}}&3&3}{1&3&3&3&3} & 1.00 \\
\addlinespace[2pt]\cmidrule(l{1pt}r{1pt}){1-4}\cmidrule(l{1pt}r{1pt}){5-8}
G2-C19 & saf, cro & \ktgrid{1}{4&4&4&4&4}{36&36&\textbf{\textit{36}}&36&36}{36&36&36&36&36} & 1.00 & G2-C20 & acc, cro & \ktgrid{2}{2&2&2&2&2}{2&10&\textbf{\textit{10}}&10&10}{2&10&10&10&10} & 1.00 \\
\addlinespace[2pt]\cmidrule(l{1pt}r{1pt}){1-4}\cmidrule(l{1pt}r{1pt}){5-8}
G2-C21 & cro, acc & \ktgrid{2}{2&6&6&6&6}{2&6&\textbf{\textit{11}}&11&11}{2&6&11&11&11} & 1.00 & G2-C22 & cro, saf & \ktgrid{2}{2&14&14&14&14}{2&79&\textbf{\textit{79}}&79&79}{2&79&79&79&79} & 1.00 \\
\addlinespace[2pt]\cmidrule(l{1pt}r{1pt}){1-4}\cmidrule(l{1pt}r{1pt}){5-8}
G2-C23 & cro, saf & \ktgrid{1}{6&6&6&6&6}{6&6&\textbf{\textit{46}}&46&46}{6&6&46&46&46} & 1.00 & G2-C24 & cro, saf & \ktgrid{1}{6&6&6&6&6}{9&9&\textbf{\textit{9}}&9&9}{9&9&9&9&9} & 1.00 \\
\addlinespace[2pt]\cmidrule(l{1pt}r{1pt}){1-4}\cmidrule(l{1pt}r{1pt}){5-8}
G2-C25 & saf, sig & \ktgrid{1}{4&4&4&4&4}{4&4&\textbf{\textit{4}}&4&4}{4&4&4&4&4} & 1.00 & G2-C26 & cro, saf & \ktgrid{1}{2&2&2&2&2}{2&2&\textbf{\textit{14}}&14&14}{2&2&14&14&14} & 1.00 \\
\addlinespace[2pt]\cmidrule(l{1pt}r{1pt}){1-4}\cmidrule(l{1pt}r{1pt}){5-8}
G2-C27 & cro, sig & \ktgrid{2}{2&5&5&5&5}{2&12&\textbf{\textit{12}}&12&12}{2&12&12&12&12} & 1.00 & G2-C28 & saf, acc & \ktgrid{1}{3&3&3&3&3}{3&3&\textbf{\textit{3}}&3&3}{3&3&3&3&3} & 1.00 \\
\addlinespace[2pt]\cmidrule(l{1pt}r{1pt}){1-4}\cmidrule(l{1pt}r{1pt}){5-8}
G2-C29 & saf, acc & \ktgrid{1}{8&8&8&8&8}{8&18&\textbf{\textit{18}}&18&18}{8&18&18&18&18} & 1.00 & G2-C30 & acc, cro & \ktgrid{2}{2&2&2&2&2}{2&2&\textbf{\textit{7}}&7&7}{2&2&7&7&7} & 1.00 \\
\addlinespace[2pt]\cmidrule(l{1pt}r{1pt}){1-4}\cmidrule(l{1pt}r{1pt}){5-8}
G2-C31 & saf, cro & \ktgrid{1}{2&2&2&2&2}{5&5&\textbf{\textit{5}}&5&5}{5&5&5&5&5} & 1.00 &  & & &  \\
\addlinespace[2pt]
\end{longtable}
}

\providecommand{\ktgrid}[4]{{\scriptsize\setlength{\tabcolsep}{1.5pt}\renewcommand{\arraystretch}{0.85}\begin{tabular}{@{}c|ccccc@{}}$K\backslash T$ & $0.8$ & $0.7$ & $0.6$ & $0.5$ & $0.4$ \\ \cmidrule(lr){1-6}$0$ & \multicolumn{5}{c}{#1} \\ $1$ & #2 \\ $2$ & #3 \\ $3$ & #4 \\ \end{tabular}}}

{\footnotesize\setlength{\tabcolsep}{3pt}
\begin{longtable}{@{}l l c c @{\hspace{4pt}}|@{\hspace{4pt}} l l c c@{}}
\caption{Per-case $K\times T$ Pareto grid, \textbf{1-active} group (17 cases). Layout and metrics as in Table~\ref{tab:kt-grid-3active}; with one active preference $K{=}1,2,3$ coincide.}\label{tab:kt-grid-1active}\\
\toprule
Case & Active prefs & $|P|$ & $\pi_D$ & Case & Active prefs & $|P|$ & $\pi_D$ \\
\midrule \endfirsthead
\multicolumn{8}{@{}l@{}}{\scriptsize\emph{Table~\ref{tab:kt-grid-1active} (continued).} Each $|P|$ cell is a $4$-$K\times5$-$T$ grid.}\\[2pt]
\toprule
Case & Active prefs & $|P|$ & $\pi_D$ & Case & Active prefs & $|P|$ & $\pi_D$ \\
\midrule \endhead
\midrule \multicolumn{8}{r@{}}{\footnotesize continued on next page}\\ \endfoot
\bottomrule \endlastfoot
G1-C1 & saf & \ktgrid{2}{4&4&4&4&4}{4&4&\textbf{\textit{4}}&4&4}{4&4&4&4&4} & 1.00 & G1-C2 & cro & \ktgrid{1}{10&10&10&10&10}{10&10&\textbf{\textit{10}}&10&10}{10&10&10&10&10} & 1.00 \\
\addlinespace[2pt]\cmidrule(l{1pt}r{1pt}){1-4}\cmidrule(l{1pt}r{1pt}){5-8}
G1-C3 & sig & \ktgrid{1}{4&4&4&4&4}{4&4&\textbf{\textit{4}}&4&4}{4&4&4&4&4} & 1.00 & G1-C4 & saf & \ktgrid{2}{3&3&3&3&3}{3&3&\textbf{\textit{3}}&3&3}{3&3&3&3&3} & 1.00 \\
\addlinespace[2pt]\cmidrule(l{1pt}r{1pt}){1-4}\cmidrule(l{1pt}r{1pt}){5-8}
G1-C5 & cro & \ktgrid{1}{1&2&2&2&2}{1&2&\textbf{\textit{2}}&2&2}{1&2&2&2&2} & 1.00 & G1-C6 & cro & \ktgrid{1}{2&2&2&2&2}{2&2&\textbf{\textit{2}}&2&2}{2&2&2&2&2} & 1.00 \\
\addlinespace[2pt]\cmidrule(l{1pt}r{1pt}){1-4}\cmidrule(l{1pt}r{1pt}){5-8}
G1-C7 & cro & \ktgrid{1}{1&4&4&4&4}{1&4&\textbf{\textit{4}}&4&4}{1&4&4&4&4} & 1.00 & G1-C8 & saf & \ktgrid{1}{4&4&4&4&4}{4&4&\textbf{\textit{4}}&4&4}{4&4&4&4&4} & 1.00 \\
\addlinespace[2pt]\cmidrule(l{1pt}r{1pt}){1-4}\cmidrule(l{1pt}r{1pt}){5-8}
G1-C9 & saf & \ktgrid{1}{4&4&4&4&4}{4&4&\textbf{\textit{4}}&4&4}{4&4&4&4&4} & 1.00 & G1-C10 & cro & \ktgrid{1}{1&5&5&5&5}{1&5&\textbf{\textit{5}}&5&5}{1&5&5&5&5} & 1.00 \\
\addlinespace[2pt]\cmidrule(l{1pt}r{1pt}){1-4}\cmidrule(l{1pt}r{1pt}){5-8}
G1-C11 & cro & \ktgrid{1}{5&5&5&5&5}{5&5&\textbf{\textit{5}}&5&5}{5&5&5&5&5} & 1.00 & G1-C12 & acc & \ktgrid{1}{1&1&5&5&5}{1&1&\textbf{\textit{5}}&5&5}{1&1&5&5&5} & 1.00 \\
\addlinespace[2pt]\cmidrule(l{1pt}r{1pt}){1-4}\cmidrule(l{1pt}r{1pt}){5-8}
G1-C13 & cro & \ktgrid{1}{1&8&8&8&8}{1&8&\textbf{\textit{8}}&8&8}{1&8&8&8&8} & 1.00 & G1-C14 & cro & \ktgrid{1}{1&8&8&8&8}{1&8&\textbf{\textit{8}}&8&8}{1&8&8&8&8} & 1.00 \\
\addlinespace[2pt]\cmidrule(l{1pt}r{1pt}){1-4}\cmidrule(l{1pt}r{1pt}){5-8}
G1-C15 & saf & \ktgrid{1}{4&4&4&4&4}{4&4&\textbf{\textit{4}}&4&4}{4&4&4&4&4} & 1.00 & G1-C16 & saf & \ktgrid{1}{6&6&6&6&6}{6&6&\textbf{\textit{6}}&6&6}{6&6&6&6&6} & 1.00 \\
\addlinespace[2pt]\cmidrule(l{1pt}r{1pt}){1-4}\cmidrule(l{1pt}r{1pt}){5-8}
G1-C17 & cro & \ktgrid{1}{4&4&4&4&4}{4&4&\textbf{\textit{4}}&4&4}{4&4&4&4&4} & 1.00 &  & & &  \\
\addlinespace[2pt]
\end{longtable}
}

\providecommand{\ktgrid}[4]{{\scriptsize\setlength{\tabcolsep}{1.5pt}\renewcommand{\arraystretch}{0.85}\begin{tabular}{@{}c|ccccc@{}}$K\backslash T$ & $0.8$ & $0.7$ & $0.6$ & $0.5$ & $0.4$ \\ \cmidrule(lr){1-6}$0$ & \multicolumn{5}{c}{#1} \\ $1$ & #2 \\ $2$ & #3 \\ $3$ & #4 \\ \end{tabular}}}

{\footnotesize\setlength{\tabcolsep}{3pt}
\begin{longtable}{@{}l l c c @{\hspace{4pt}}|@{\hspace{4pt}} l l c c@{}}
\caption{Per-case $K\times T$ Pareto grid, \textbf{0-active} group (5 cases). Layout and metrics as in Table~\ref{tab:kt-grid-3active}; no preference ever clears $T$, so every setting returns the baseline.}\label{tab:kt-grid-0active}\\
\toprule
Case & Active prefs & $|P|$ & $\pi_D$ & Case & Active prefs & $|P|$ & $\pi_D$ \\
\midrule \endfirsthead
\multicolumn{8}{@{}l@{}}{\scriptsize\emph{Table~\ref{tab:kt-grid-0active} (continued).} Each $|P|$ cell is a $4$-$K\times5$-$T$ grid.}\\[2pt]
\toprule
Case & Active prefs & $|P|$ & $\pi_D$ & Case & Active prefs & $|P|$ & $\pi_D$ \\
\midrule \endhead
\midrule \multicolumn{8}{r@{}}{\footnotesize continued on next page}\\ \endfoot
\bottomrule \endlastfoot
G0-C1 & --- & \ktgrid{2}{2&2&2&2&2}{2&2&\textbf{\textit{2}}&2&2}{2&2&2&2&2} & 1.00 & G0-C2 & --- & \ktgrid{2}{2&2&2&2&2}{2&2&\textbf{\textit{2}}&2&2}{2&2&2&2&2} & 1.00 \\
\addlinespace[2pt]\cmidrule(l{1pt}r{1pt}){1-4}\cmidrule(l{1pt}r{1pt}){5-8}
G0-C3 & --- & \ktgrid{2}{2&2&2&2&2}{2&2&\textbf{\textit{2}}&2&2}{2&2&2&2&2} & 1.00 & G0-C4 & --- & \ktgrid{1}{1&1&1&1&1}{1&1&\textbf{\textit{1}}&1&1}{1&1&1&1&1} & 1.00 \\
\addlinespace[2pt]\cmidrule(l{1pt}r{1pt}){1-4}\cmidrule(l{1pt}r{1pt}){5-8}
G0-C5 & --- & \ktgrid{1}{1&1&1&1&1}{1&1&\textbf{\textit{1}}&1&1}{1&1&1&1&1} & 1.00 &  & & &  \\
\addlinespace[2pt]
\end{longtable}
}

\section{Curated Keyword Dictionary for Baseline Classifier}
\label{apx:keyword-lists}

This section details the complete sets of curated keywords utilized by the rule-based baseline system (described in Section~\ref{sec:keyword-baseline}) to detect user travel preferences from natural language requests. The dictionary consists of 126 unique natural language triggers and semantic proxies, categorized into four core transportation preference dimensions. The keywords for each dimension are presented below in alphabetical order:

\begin{itemize}
    \item \textbf{Accessibility (25 keywords):} 
    access needs, accessible, accessible platform, assistance, barrier-free, carrying, disabled, easy access, elevator, heavy bags, lift, luggage, mobility, mobility equipment, physical, platform, ramp, stairs, step free, step-free, stroller, suitcase, walking, walking distance, wheelchair.

    \item \textbf{Safety (31 keywords):} 
    alone, area safety, avoid, bright, busy area, cctv, crime, dangerous, dark, emergency, feel secure, gangster, late night, lighting, main road, monitored, neighborhood, night time, personal security, police, populated, risk, safe, safety, secure, security, threat, unsafe, visible, well lit, well-lit.

    \item \textbf{Crowdedness (31 keywords):} 
    anxiety, avoid crowds, breathing room, busy, claustrophobic, comfortable, congested, crowd, crowded, empty, full, jam, less busy, less crowded, not busy, overwhelming, packed, peak hour, personal space, quiet, room, rush, rush hour, sardines, seat, seating, space, spacious, standing, stress, uncrowded.

    \item \textbf{Sightseeing (39 keywords):} 
    above ground, architecture, art, attraction, attractions, beautiful, bridge, cathedral, church, cultural, culture, enjoyable, experience, explore, exploring, gallery, garden, historic, holiday, interesting, joy, landmarks, leisure, monument, museum, overground, palace, park, river, scenic, sights, sightseeing, street view, tourism, tourist, vacation, view, views, window.
\end{itemize}

\section{Multi-turn User Request Parsing}
\label{apx:multi-turn}
The evaluations in Section~\ref{sec:exp1} assessed the parsing component on single-turn requests from the User Study Test Set. In practice, users may also interact with the system across multiple turns, for example, omitting key trip details initially or revising their preferences during the conversation. To evaluate parsing performance under these conditions, an additional multi-turn test set was constructed. Each request in the User Study Test Set was randomly assigned with a fixed seed to four dialogue patterns, with each base request used only once.

\subsection{Multi-turn Test Set Construction}
The multi-turn test set was created by rule-based transformations of the original single-turn requests and their structured labels. Four dialogue patterns were considered: (i) requests with the origin omitted, (ii) requests with the requested arrival time omitted, (iii) requests with both origin and requested arrival time omitted, and (iv) complete requests followed by later preference revisions. For incomplete-information cases (i-iii), the missing fields were removed from the initial request and then supplied in subsequent user turns. For the preference-revision cases (iv), the original request was preserved and followed by short, templated updates (e.g., 'I care more about safety'), and the corresponding preference score in the ground-truth output was set to 1.0. An example dialogue is shown in Figure~\ref{fig:multiturn-example}. The Multi-turn Test Set should be interpreted as a controlled test of conversational parsing under common interaction patterns, rather than a substitute for naturally collected multi-turn dialogues, which is left for future work.

\begin{figure}[h]
\centering
\begin{tcolorbox}[
  colback=white,
  colframe=gray!60,
  boxrule=0.5pt,
  arc=2pt,
  left=6pt,
  right=6pt,
  top=6pt,
  bottom=6pt,
  width=\textwidth
]
\small

\textit{System instruction (abridged):} \textit{``You are a strict information extractor for public transit requests. If start\_stop, end\_stop, requested\_arrival\_time, or preference is missing, ask one short clarification question. If all are provided, output only one JSON object ...''}

\vspace{4pt}
\hrule
\vspace{6pt}

\textbf{Turn 1} \\[4pt]
\textbf{User:} ``I want to get to Stratford from the Sainsbury's in Whitechapel by 2 pm. I am carrying groceries, so please make sure the route doesn't have crowded spaces.'' \\[4pt]
\textbf{Assistant (ChatPlanner response):} \\
\texttt{\{...\}}

\vspace{6pt}
\hrule
\vspace{6pt}

\textbf{Turn 2} \\[4pt]
\textbf{User:} ``I care more about safety.'' \\[4pt]
\textbf{Assistant (ChatPlanner response):} \\
\texttt{\{...\}}

\vspace{6pt}
\hrule
\vspace{6pt}

\textbf{Turn 3} \\[4pt]
\textbf{User:} ``Please avoid crowds more.'' \\[4pt]
\textbf{Assistant (ChatPlanner response):} \\
\texttt{\{...\}} \\
[8pt]
\textit{\textbf{Ground-truth (for evaluation):}} \\
\textit{\texttt{\{"start\_stop": "Whitechapel", "end\_stop": "Stratford",} \\
\texttt{~"requested\_arrival\_time": "14:00:00",} \\
\texttt{~"scores": \{"accessibility": 0.7, "crowdedness": 1,} \\
\texttt{~~"safety": 1, "sightseeing": 0.2\}\}}}

\end{tcolorbox}
\caption{Each dialogue instance consists of the user's messages across turns and a ground-truth structured output used for evaluation. The system instruction shown at the top is shared across all dialogues and is not part of any individual dialogue. During evaluation, the model generates its own response at each assistant turn, and the final output is compared against the ground-truth.}
\label{fig:multiturn-example}
\end{figure}

\subsection{Evaluation Setup}
Each configuration was evaluated by replaying the Multi-turn Test Set dialogue one turn at a time. The system prompt and user messages were provided sequentially, and whenever an assistant turn was reached, the LLM was asked to generate its own response based on the preceding dialogue history. The LLM reply was then added to the dialogue history before the next user turn was provided. In this way, the evaluation tests whether the LLM can ask for missing information when needed and produce the final structured output once sufficient information has been supplied. The same LLM configurations as in the single-turn experiments in Section~\ref{sec:exp1} were tested here. To maintain comparability, the RAG-based configurations used the same retrieval library as in the single-turn evaluation.

\subsection{Multi-turn Evaluation Metrics}
The same metrics as above are reported: EM\_start, EM\_end, EM\_time, EM\_all, MAE, Subset Accuracy, and Pearson correlation coefficient. In addition, ARF\_turn0 (Ask-Right-Field Accuracy at the first assistant turn) is introduced, which measures whether the LLM's first clarification question asks for the field or fields that are missing from the request.

\subsection{Results and Discussions}

\begin{table}[htbp]
\small
\centering
\caption{Multi-turn performance on the Multi-turn Test Set.}
\label{tab:multiturn-results}
\setlength{\tabcolsep}{3pt}
\renewcommand{\arraystretch}{1.05}
\begin{tabularx}{\textwidth}{L d d d d d d d d}
\toprule
\makecell{\scriptsize\textbf{Label}} &
\multicolumn{1}{c}{\makecell{\scriptsize\textbf{EM}\\\scriptsize\textbf{start}}} &
\multicolumn{1}{c}{\makecell{\scriptsize\textbf{EM}\\\scriptsize\textbf{end}}} &
\multicolumn{1}{c}{\makecell{\scriptsize\textbf{EM}\\\scriptsize\textbf{time}}} &
\multicolumn{1}{c}{\makecell{\scriptsize\textbf{EM}\\\scriptsize\textbf{all}}} &
\multicolumn{1}{c}{\makecell{\scriptsize\textbf{MAE}}} &
\multicolumn{1}{c}{\makecell{\scriptsize\textbf{Subset}\\\scriptsize\textbf{Accuracy}}} &
\multicolumn{1}{c}{\makecell{\scriptsize\textbf{Pearson}}} &
\multicolumn{1}{c}{\makecell{\scriptsize\textbf{ARF}\\\scriptsize\textbf{turn0}}} \\
\midrule
Llama Base                & 0.458 & 0.417 & 0.442 & 0.342 & 0.390 & 0.058 & 0.103 & 0.333 \\
Qwen Base                 & 0.358 & 0.342 & 0.417 & 0.258 & 0.398 & 0.017 & 0.133 & 0.433 \\
Llama Base + RAG          & 0.617 & 0.508 & 0.642 & 0.417 & 0.270 & 0.342 & 0.506 & 0.500 \\
Qwen Base + RAG           & 0.642 & 0.583 & 0.683 & 0.542 & 0.278 & 0.317 & 0.439 & 0.500 \\
Llama Fine-tuned          & 0.642 & 0.592 & 0.700 & 0.500 & 0.265 & 0.400 & 0.511 & 0.617 \\
Qwen Fine-tuned           & 0.658 & 0.617 & 0.725 & 0.567 & 0.274 & 0.392 & 0.451 & 0.650 \\
Llama Fine-tuned + RAG    & 0.742 & \textbf{0.750} & 0.833 & 0.658 & 0.226 & 0.583 & \textbf{0.539} & 0.733 \\
Qwen Fine-tuned + RAG     & \textbf{0.758} & 0.733 & \textbf{0.858} & \textbf{0.675} & \textbf{0.219} & \textbf{0.592} & 0.513 & \textbf{0.767} \\
\bottomrule
\end{tabularx}
\end{table}

The results in Table~\ref{tab:multiturn-results} show that the same performance hierarchy observed in the single-turn experiments holds in the multi-turn setting. Fine-tuning combined with RAG consistently outperforms all other configurations across all metrics. For ARF\_turn0, the best configuration correctly identifies the missing field(s) in more than 75\% of cases, indicating the ChatPlanner's configuration can often recognize what information is lacking, while there is clear room for improvement. Overall, performance is lower than in the single-turn setting, which is expected given the added complexity of handling clarification and preference revision across multiple turns, but the results still demonstrate that the parsing component can function in a conversational setting. This performance could be further improved through several potential approaches, such as incorporating multi-turn dialogue examples into the fine-tuning data, collecting and training on real user conversation dialogues from a deployed system, and so on.

\end{document}